\pdfoutput=1
\documentclass[10pt,twocolumn,letterpaper]{article}

\newif\ifarxiv
\arxivtrue

\ifarxiv
\usepackage[pagenumbers]{iccv} %
\usepackage[absolute,overlay]{textpos}
\else
\usepackage{iccv}              %
\fi

\def\MYTITLE{Simultaneous Motion And Noise Estimation with Event Cameras}

\definecolor{iccvblue}{rgb}{0.21,0.49,0.74}
\ifarxiv
\usepackage[breaklinks,colorlinks,allcolors=iccvblue]{hyperref}
\hypersetup{
  pdftitle={\MYTITLE},
  pdfsubject={Computer Vision, Robotics, Deep Learning},
  pdfauthor={Shintaro Shiba, Yoshimitsu Aoki, Guillermo Gallego},
  pdfkeywords={Event Cameras, Motion Estimation, Denoising}
}
\else
\usepackage[breaklinks,colorlinks,allcolors=iccvblue]{hyperref}
\fi

\usepackage{amsmath,amssymb,amsfonts}
\usepackage{graphicx}
\usepackage{caption}
\usepackage{subcaption}
\usepackage{makecell}
\usepackage{booktabs}
\usepackage{siunitx}
\usepackage{adjustbox} %
\usepackage{array}
\usepackage{multirow}
\usepackage{rotating}
\usepackage{etoolbox}
\robustify\bfseries

\usepackage{url}
\usepackage{color}

\def\tref{t_\text{ref}} %
\def\pol{p} %

\def\cE{\mathcal{E}} %
\def\numEvents{N_e} %
\def\numPixels{N_p} %
\def\Warp{\mathbf{W}}

\def\bx{\mathbf{x}}
\def\bparams{\btheta}
\def\Rot{\mathtt{R}}

\def\pol{p}
\def\velflow{\mathbf{v}}
\def\angvel{\boldsymbol{\omega}} %

\def\cN{\mathcal{N}} %
\def\bmu{\boldsymbol{\mu}} %

\def\IWE{I} %

\def\mId{\mathtt{Id}} %
\def\variance{\operatorname{Var}}

\newcommand{\unumr}[2]{\multicolumn{1}{r}{\underline{\tablenum[table-format={#1}]{#2}}}}  %
\newcommand{\bnum}[1]{\bfseries #1}

\definecolor{light-gray}{gray}{0.6}
\newcommand\gframe[1]{{\color{light-gray}\frame{#1}}}

\usepackage[capitalize]{cleveref}
\crefname{section}{Sec.}{Secs.}
\Crefname{section}{Section}{Sections}
\Crefname{table}{Table}{Tables}
\crefname{table}{Tab.}{Tabs.}

\def\tref{t_\text{ref}} %
\def\pol{p} %

\def\cE{\mathcal{E}} %
\def\numEvents{N_e} %
\def\numPixels{N_p} %
\def\Warp{\mathbf{W}}

\def\bx{\mathbf{x}}
\def\bparams{\boldsymbol{\theta}}
\def\Rot{\mathtt{R}}

\def\pol{p}
\def\velflow{\mathbf{v}}
\def\angvel{\boldsymbol{\omega}} %
\def\bphi{\boldsymbol{\phi}}
\def\bmu{\boldsymbol{\mu}}
\def\mId{\mathtt{Id}}
\def\IWE{I}

\def\cN{\mathcal{N}} %
\def\variance{\operatorname{Var}}

\def\snratio{\tau}

\def\cEsig{\cE_\text{signal}}
\def\cEnoi{\cE_\text{noise}}

\newcommand{\gblue}[1]{\textcolor{black}{#1}} %

\usepackage[super]{nth}

\title{\MYTITLE}

\author{Shintaro Shiba$^{1,2,3}$, Yoshimitsu Aoki$^{1}$ and Guillermo Gallego$^{3,4}$\\
$^{1}$~Keio University,
$^{2}$~Woven by Toyota, Inc., Japan.
$^{3}$~Technische Universit\"at Berlin,
$^{4}$~Einstein Center \\Digital Future, Robotics Institute Germany, and Science of Intelligence Excellence Cluster, Germany.
}

\begin{document}
\maketitle

\ifarxiv
\definecolor{somegray}{gray}{0.5}
\newcommand{\darkgrayed}[1]{\textcolor{somegray}{#1}}
\begin{textblock}{11}(2.5, 0.6)
\begin{center}
\darkgrayed{This paper has been accepted for publication at the\\
IEEE International Conference on Computer Vision (ICCV), Honolulu, 2025.
\copyright IEEE}
\end{center}
\end{textblock}
\fi

\begin{abstract}

Event cameras are emerging vision sensors whose noise is challenging to characterize.
Existing denoising methods for event cameras are often designed in isolation and thus consider other tasks, such as motion estimation, separately (i.e., sequentially after denoising).
However, motion is an intrinsic part of event data, since scene edges cannot be sensed without motion.
We propose, to the best of our knowledge, the first method that simultaneously estimates motion in its various forms (e.g., ego-motion, optical flow) and noise.
The method is flexible, as it allows replacing the one-step motion estimation of
the widely-used Contrast Maximization framework with any other motion estimator,
such as deep neural networks.
The experiments show that the proposed method achieves state-of-the-art results on the E-MLB denoising benchmark and competitive results on the DND21 benchmark,
while demonstrating effectiveness across motion estimation and intensity reconstruction tasks.
Our approach advances event-data denoising theory and expands practical denoising use-cases via open-source code.
Project page: \url{https://github.com/tub-rip/ESMD}
\end{abstract}

\section{Introduction}
\label{sec:intro}

Event cameras are emerging vision sensors that 
overcome shortcomings of conventional cameras (e.g., motion blur, limited dynamic range, data and power efficiency),
yet they suffer from a considerable amount of noise due to their novelty and operation in low-power (transistor subthreshold) conditions \cite{Gallego20pami,Lichtsteiner08ssc}.
Since event cameras are suitable for many computer vision tasks, especially motion-related tasks,
it is paramount to classify event data that is related to motion (i.e., \emph{signal}) and that is not (i.e., \emph{noise}).
However, denoising event data is a challenging problem because noise properties have yet to be fully characterized \cite{Guo22pami,Graca25iisw} and it is infeasible to define ground-truth (GT) noise labels
in real-world data recordings.
Previous works either
leveraged simulation
or prepared real event data and aggressively suppressed noise occurrence.

\begin{figure}[t]
  \centering
  {{\includegraphics[clip,trim={0.5cm 2.2cm 9.8cm 0.5cm},width=\linewidth]{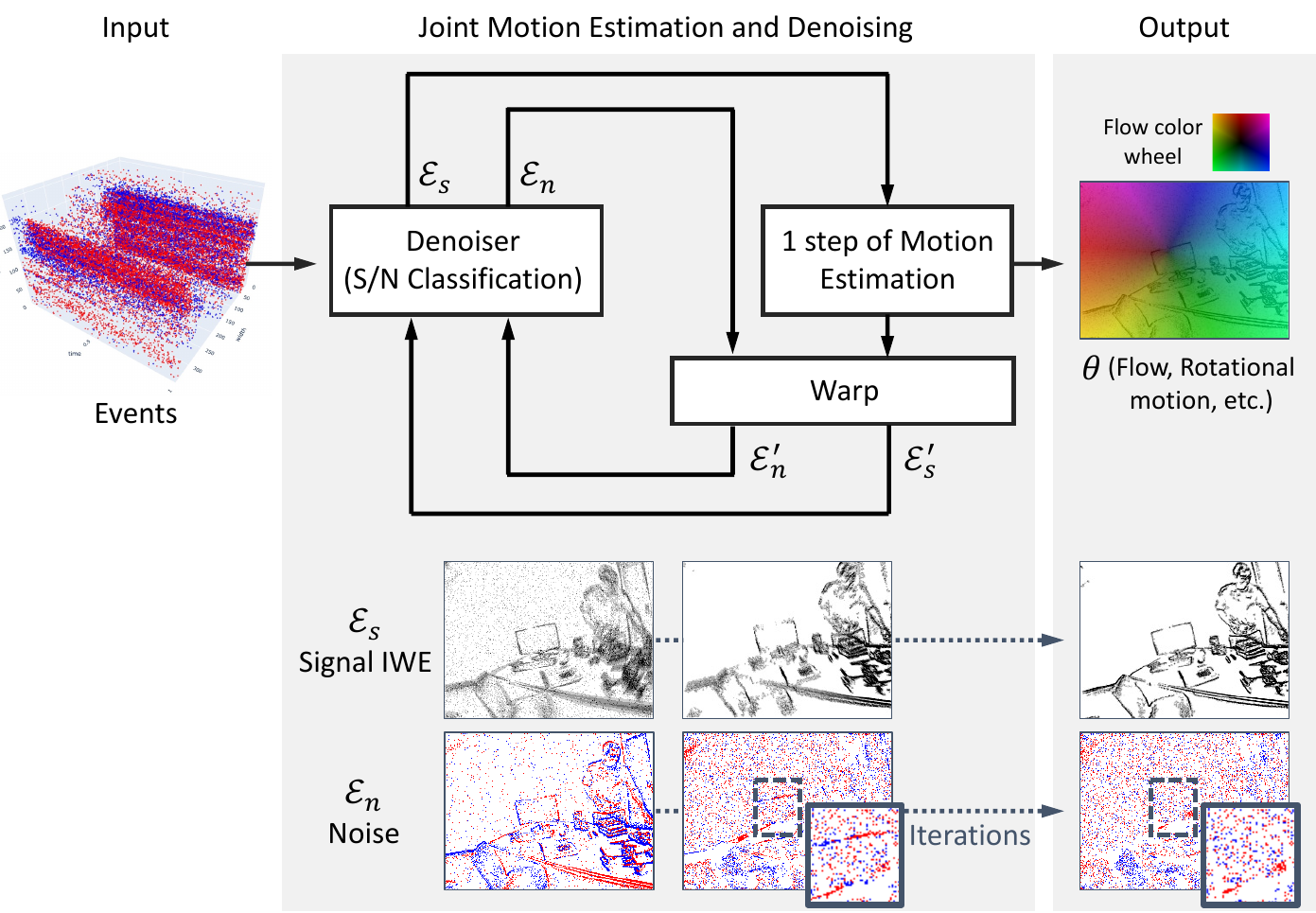}}}
  \vspace{-4.5ex}
\caption{Overview of the proposed method. We use only raw events as input, and estimate signal events, noise events, and motion (e.g., dense optical flow and ego-motion estimation).}
\label{fig:eyeCatcher}
\vspace{-1ex}
\end{figure}

However, such artificial modifications on the real event data could change event signal/noise characteristics, possibly deteriorating the information that the original event had for downstream tasks (e.g., motion estimation).
Our work is inspired by the observation that
motion is an intrinsic part of event data since scene edges cannot be sensed without motion. 
Rather than solving denoising and motion estimation tasks separately,
our idea is to leverage motion information to improve denoising (as ``vice-versa'' is well known).
In this work (\cref{fig:eyeCatcher}), we rethink and investigate the connection between denoising and motion estimation,
by asking ourselves whether they can be combined in a more integrated way to benefit the results of both tasks simultaneously.
We formulate the simultaneous estimation problem by leveraging the idea of motion compensation in the Contrast Maximization (CMax) framework,
defining novel metrics about the contribution to the contrast score for each event.
In summary, we present several distinctive contributions:
\begin{enumerate}
\item We propose a novel method to jointly estimate noise and motion only from event data (\cref{sec:method}).
\item We demonstrate that our joint estimation method 
($i$) achieves state-of-the-art results in existing denoising benchmarks (\cref{sec:experim:denoise}), 
and ($ii$) improves the robustness of motion estimation (CMax framework) (\cref{sec:experim:angVel}).
\item We show the flexibility of our method on ($i$) further applications, such as intensity reconstruction, validating its image quality (\cref{sec:experim:intensityRecon,sec:sensitivity}), and ($ii$) in combination with learning-based motion estimators (\cref{sec:experim:dnn}).
\end{enumerate}
We hope our work opens a new line of research for event cameras, 
by considering the problems of motion estimation and denoising together to take advantage of their coaction.

\section{Related Work}
\label{sec:related}

Denoising is a fundamental problem in event cameras
since there are a lot of leakage events \cite{Guo22pami},
and event-processing algorithms may suffer in scenarios with high noise \cite{Arja23cvprw}.
Removing event noise has a double effect: it reduces the load of subsequent processing stages (as the load is proportional to the number of events) and improves their output (as the input data becomes of higher quality).
In this work, we focus on
noise due to the leakage inside the camera hardware, such as background activity (BA) events \cite{Lichtsteiner08ssc}.
Events due to flickering or active lights
are therefore not our priority.

Classical denoising works focus on the spatio-temporal correlation of events,
as BA events have fewer neighboring events in space-time coordinates than signal events.
This approach includes widely-used BA Filter (BAF) \cite{Delbruck08issle},
and other types of spatio-temporal filters \cite{Liu15iscas,Khodamoradi18tetc,Wu20tm,Guo20aspdac}.
Some works utilize additional sensor inputs (e.g., frames) to estimate noise probabilities \cite{Baldwin20cvpr,Duan21cvpr,Wang20cvpr,Karmokar25wacv,Zhang24tip},
since edges are informative of the signal events produced.

Recently,
learning-based approaches have gained more attention, such as
multilayer perceptrons \cite{Guo22pami},
convolutional neural networks \cite{Baldwin20cvpr,Duan21cvpr,Afshar20jsen},
graph neural networks \cite{Alkendi22tnnls},
and transformers \cite{Jiang24eccv}.
They provide the probability of noise (i.e., signal-noise classification).
Nonetheless, the challenge of learning-based approaches is the need for ground truth (GT) noise labels for training.
GT labels have been obtained through simulation (e.g., \cite{Fang22aednet,Guo22pami}) as well as real-world recordings,
however, real-world recordings aggressively remove noisy events using existing filters (e.g., \cite{Guo22pami})
or use unnatural illumination settings (e.g., \cite{Jiang24eccv}),
by leveraging the differences of noise occurrence due to different scene brightness \cite{Guo22pami,Lichtsteiner06isscc}.
While these approaches are effective for benchmarking,
the challenge of obtaining GT labels for practical real-world datasets remains unsolved.

Our work builds upon a series of successful methods for motion estimation and their applications.
Event-based motion estimation is an important topic since event cameras naturally respond to motion in the scenes \cite{Gallego20pami}.
In particular, Contrast Maximization (CMax) \cite{Gallego18cvpr,Shiba24pami} is a state-of-the-art framework for motion estimation,
which has various applications, 
such as rotational motion estimation \cite{Gallego17ral,Gu21iccv,Peng21pami,Nunes21pami,Kim21ral},
optical flow estimation \cite{Zhu19cvpr,Paredes23iccv,Shiba22eccv,Hamann24eccv,Guo25iccv},
and motion segmentation \cite{Stoffregen19iccv,Zhou21tnnls,Yamaki25cvprw}.
We investigate the connection and potential benefits of applying the CMax framework to denoising during the estimation process (i.e., simultaneously), 
demonstrating motion estimation improvements as well as competitive denoising accuracy.

A prior work, ``ProgressiveMotionSeg'' \cite{Chen22aaai},
has extended the CMax-based segmentation method \cite{Stoffregen19iccv} with extra weights that accounted for how much each event contributed to the estimated motion clusters.
However, \cite{Chen22aaai} mainly focuses on the task of motion segmentation using low degrees-of-freedom (DOF) warps (e.g., 2-DOF feature flow) and does not discuss denoising efficacy or evaluate on standard denoising benchmarks.
In contrast, our method supports various types of motion estimation (both low-DOF and high-DOF warps, such as optical flow), 
and considers that noise shall not be used to estimate motion, as from first principles we deem noise and motion to be uncorrelated.

\section{Methodology}
\label{sec:method}

\textbf{Event cameras} %
consist of independent pixels that generate
asynchronous ``events''
when the logarithmic brightness at the pixel increases or decreases by a predefined contrast sensitivity.
Each event $e_k \doteq (\bx_k,t_k,\pol_k)$ contains the pixel coordinates $\bx_k$, the timestamp $t_k$,
and polarity $\pol_k \in \{+1,-1\}$ of the brightness change. 
Events occur asynchronously and sparsely on the pixel grid, resulting in a variable data rate based on the scene texture and dynamics.

\subsection{Contrast Maximization}
\label{sec:method:cmax}

The CMax framework \cite{Gallego18cvpr} assumes events $\cE \doteq \{e_k\}_{k=1}^{\numEvents}$
are caused by moving edges, %
and it transforms their coordinates %
according to a motion model $\Warp$,
producing a set of warped events $\cE'_{\tref} \doteq \{e'_k\}_{k=1}^{\numEvents}$ at a reference time $\tref$:
\begin{equation}
\vspace{-0.6ex}
e_k \doteq (\bx_k,t_k,\pol_k) \;\,\mapsto\;\, 
e'_k \doteq (\bx'_k,\tref,\pol_k).
\vspace{-0.2ex}
\end{equation}
The warp $\bx'_k = \Warp(\bx_k,t_k; \bparams)$ transports each event from $t_k$ to $\tref$ along the motion curve that passes through it. 
Then, they are aggregated on an image of warped events (IWE):
\begin{equation}
\label{eq:IWE}
\vspace{-0.6ex}
\textstyle
\IWE(\bx; \cE'_{\tref}, \bparams) \doteq \sum_{k=1}^{\numEvents} \delta (\bx - \bx'_k),
\vspace{-0.2ex}
\end{equation}
where each pixel $\bx$ sums the number of warped events $\bx'_k$ that fall within it.
The Dirac delta is approximated by a Gaussian, 
$\delta(\bx-\bmu)\approx\cN(\bx;\bmu,\epsilon^2\mId)$ with $\epsilon=1$ pixel.

\def\textWidth{0.03\linewidth}
\def\figWidth{0.15\linewidth} %
\begin{figure*}[t]
	\centering
    {\scriptsize
    \setlength{\tabcolsep}{1pt}
	\begin{tabular}{
	>{\centering\arraybackslash}m{\textWidth}
	>{\centering\arraybackslash}m{\figWidth} 
	>{\centering\arraybackslash}m{\figWidth} 
	>{\centering\arraybackslash}m{\figWidth} 
	>{\centering\arraybackslash}m{\figWidth} 
	>{\centering\arraybackslash}m{\figWidth}
	>{\centering\arraybackslash}m{\figWidth}}
		\\

            \rotatebox{90}{\makecell{GasTank (E-MLB)}}
		& \gframe{\includegraphics[width=\linewidth]{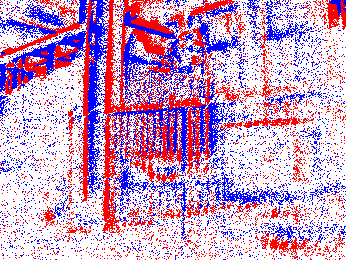}}
		&\gframe{\includegraphics[width=\linewidth]{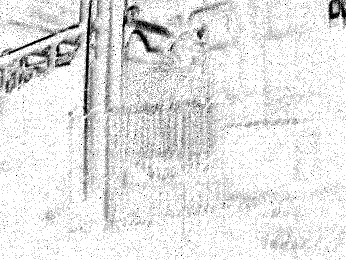}}
		&\gframe{\includegraphics[width=\linewidth]{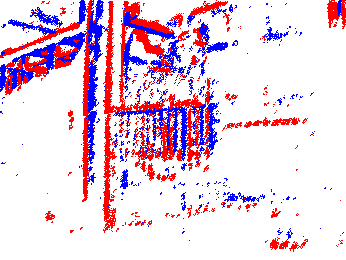}}
		&\gframe{\includegraphics[width=\linewidth]{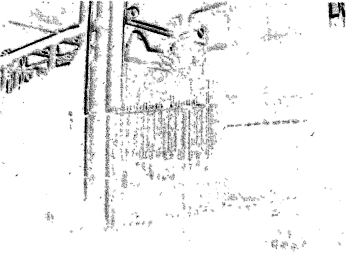}}
		&\gframe{\includegraphics[width=\linewidth]{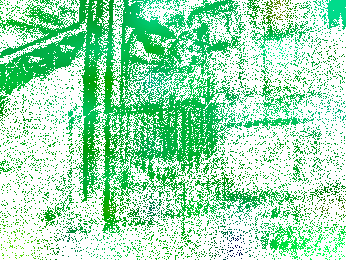}}
		&\gframe{\includegraphics[width=\linewidth]{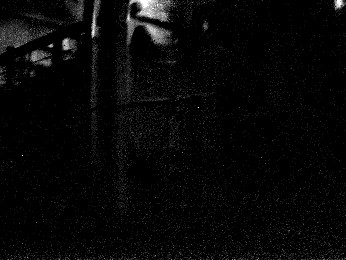}}
		\\

              \rotatebox{90}{\makecell{LEGO (E-MLB)}}
		& \gframe{\includegraphics[width=\linewidth]{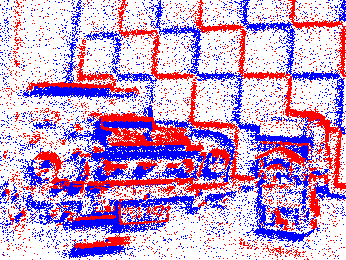}}
		&\gframe{\includegraphics[width=\linewidth]{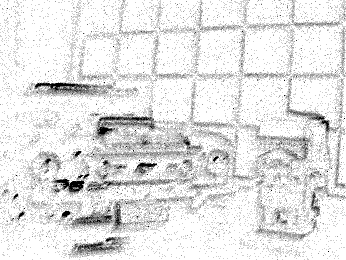}}
		&\gframe{\includegraphics[width=\linewidth]{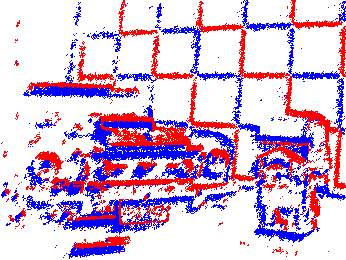}}
		&\gframe{\includegraphics[width=\linewidth]{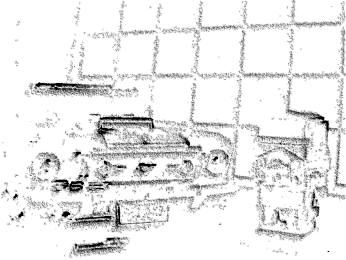}}
		&\gframe{\includegraphics[width=\linewidth]{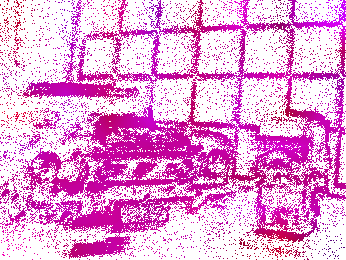}}
		&\gframe{\includegraphics[width=\linewidth]{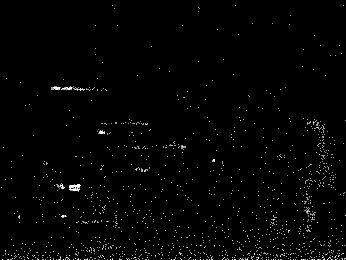}}
		\\
  
            \rotatebox{90}{\makecell{Beam (E-MLB)}}
		& \gframe{\includegraphics[width=\linewidth]{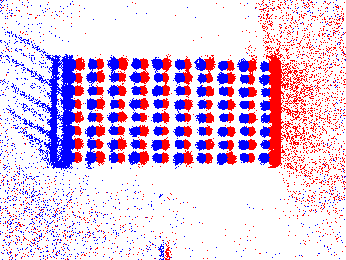}}
		&\gframe{\includegraphics[width=\linewidth]{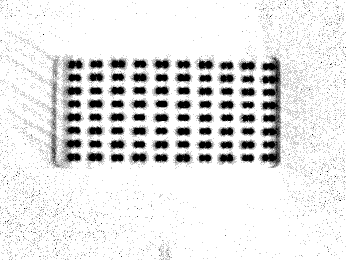}}
		&\gframe{\includegraphics[width=\linewidth]{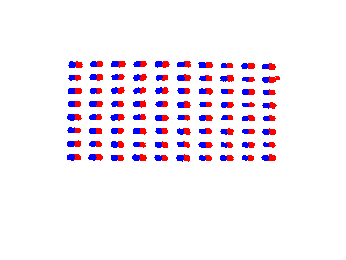}}
		&\gframe{\includegraphics[width=\linewidth]{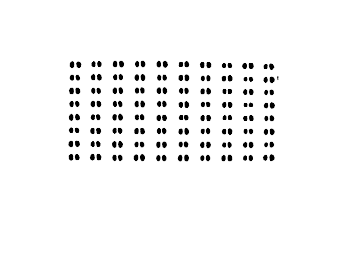}}
		&\gframe{\includegraphics[width=\linewidth]{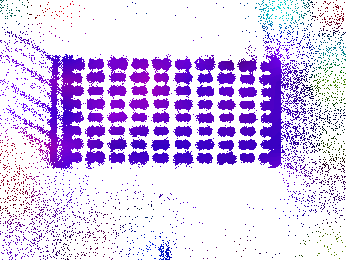}}
		&\gframe{\includegraphics[width=\linewidth]{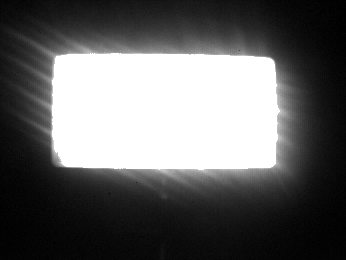}}
            \\

            \rotatebox{90}{\makecell{Driving (DND21)}}
		& \gframe{\includegraphics[width=\linewidth]{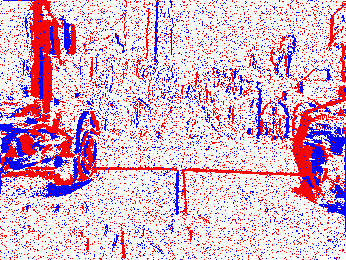}}
		&\gframe{\includegraphics[width=\linewidth]{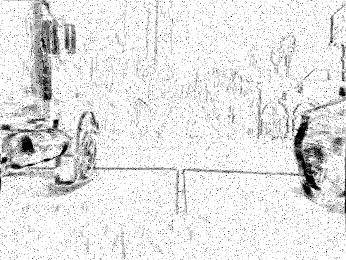}}
		&\gframe{\includegraphics[width=\linewidth]{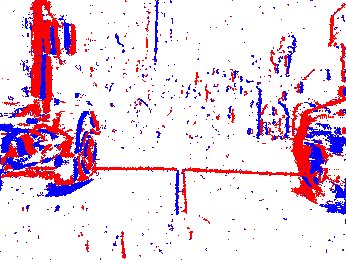}}
		&\gframe{\includegraphics[width=\linewidth]{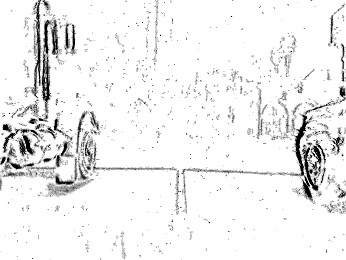}}
		&\gframe{\includegraphics[width=\linewidth]{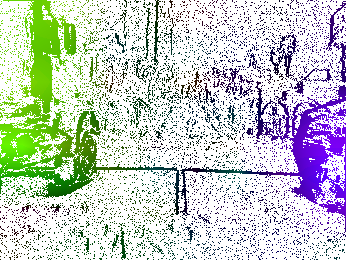}}
		&\gframe{\includegraphics[width=\linewidth]{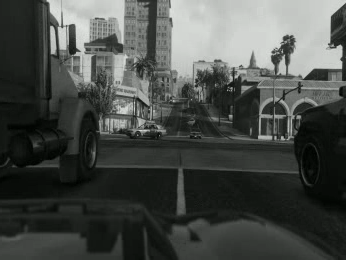}}
		\\
 
		& (a) Input (Raw Events)
		& (b) IWE (No Warp)
		& (c) Output (Signal Events)
		& (d) IWE (Signal Events)
		& (e) Estimated Flow 
		& (f) Reference Frame
		\\
	\end{tabular}
	}
	\caption{Qualitative result of denoising and flow estimation on E-MLB and DND21.
    The driving sequence has $5$-Hz artificial noise.
    (b) is an alternative visualization of (a) with the same grayscale coding as (d).
    Flow is visualized on all events and flow color is given in \cref{fig:eyeCatcher}.}
    \label{fig:result:denoiseQuality}
\end{figure*}

Next, an objective function, such as the contrast of the IWE~\eqref{eq:IWE} (image variance \cite{Gallego19cvpr}
$\variance\bigl(\IWE(\bx;\bparams)\bigr) 
\doteq \frac{1}{|\Omega|} \int_{\Omega} (\IWE(\bx;\bparams)-\mu_{\IWE})^2 d\bx$), 
measures the goodness of fit between the events and the candidate motion curves (warp).

Finally, an optimization algorithm iterates the above steps to find the motion parameters $\bparams^\ast$
that maximize the alignment of events caused by the same scene edge.
Event alignment is measured by the strength of the edges of the IWE, 
which is related to image contrast \cite{Gallego19cvpr}.

\textbf{Motion Models}. As exemplary warps $\Warp$ we focus on two popular problem settings:
rotational motion (3-DOF angular velocity estimation \cite{Peng21pami,Nunes21pami,Gu21iccv,Guo24tro}) 
and dense optical flow (pixel-wise velocity estimation \cite{Zhu19cvpr,Paredes21cvpr,Shiba22eccv,Shiba24pami}, resulting in $2\numPixels$-DOFs, where $\numPixels$ is the number of pixels).
For the angular velocity estimation during a small time interval,
the warp is parametrized by the angular velocity $\bparams \equiv \angvel = (\omega_{x},\omega_{y},\omega_{z})^{\top}$,
and 
$\bx^{h} (t) \sim \Rot(\angvel t)\,\bx^h (0)$,
using calibrated homogeneous coordinates $\bx^h$
and exponential coordinates $\Rot(\bphi) \doteq \exp(\bphi^\wedge)$ \cite{Barfoot15book,Gallego14jmiv}.
The warp for dense optical flow is
$\bx'_k = \bx_k + (t_k-\tref) \, \velflow(\bx_k)$,
where $\bparams = \{ \velflow(\bx) \}_{\bx\in\Omega}$ is a flow field on the image plane $\Omega$.

\subsection{Denoising Based on Local Contrast}
\label{sec:method:denoise}

As \cref{fig:eyeCatcher} shows, our method is iterative. 
In each iteration, motion estimation is refined and events are classified into signal $\cE_s$ or noise $\cE_n$. %
Without labels or additional information, events rely on each other to realize this classification.
To this end, we compute a score $c_k$ for each event, rank the events according to their scores, and threshold at some value $T$.
The threshold is computed by assuming the amount or ratio of noisy events is known $\eta$, i.e., $T\equiv T(\eta)$,
which allows us to have high-level control over the denoising process.
Hence we classify events into signal (the $\snratio \doteq (1-\eta)$ percentage of events with highest scores) and noise:
\begin{equation}
\begin{split}
\label{eq:denoising}
\cE_s \equiv \cEsig & \doteq \{e_k \in \cE \;|\; c_k > T(\eta) \}, \\
\cE_n \equiv \cEnoi & \doteq \cE \backslash \cEsig \quad(\text{complement of }\cEsig).
\end{split}
\end{equation}

We studied different choices for the per-event score $c_k$ based on the agreement among warped events $\cE' = \cE'_s \cup \cE'_n$, through their IWEs, and concluded on two, as follows.
Letting $\IWE^s_k \equiv \IWE^s_k(\bx'_k), \IWE^n_k \equiv \IWE^n_k(\bx'_k)$ and $\IWE_k \equiv \IWE(\bx'_k) = \IWE^s_k + \IWE^n_k$ 
be the IWE values %
of previously classified signal events, noise events and all events, respectively,
the two choices are:
($i$) $^{1}c_k \doteq \IWE_k$, 
and ($ii$) the ratio of signal events $^{2}c_k \doteq \IWE^s_k / \IWE_k$ (probabilistic viewpoint).
In fact, both are equivalent: rule $^1 c_k > T_1$ is the same as rule $^2 c_k > T_2$ 
if $T_2 = 1 - \IWE^n_k / T_1$ and the event noise $\IWE^n_k$ is known.
We use $^1 c_k$, the number of warped events (also called local contrast \cite{Gallego19cvpr,Stoffregen19iccv}), because it is the simplest to interpret. %

The intuition is that ``signal'' events (e.g., due to motion) warped according to the true motion lie all at motion-compensated edges, and therefore produce a sharp IWE.
Noise events, produced all over the image plane, are not expected to contribute to the IWE's sharpness.
Hence, we may classify events into signal or noise according to how much they contribute to edge strength. 
The IWE \eqref{eq:IWE} provides one such measure of edge strength: 
the higher $\IWE(\bx)$, the more events are warped to that pixel location $\bx$ (i.e., more events support the same scene edge), thus producing a sharper IWE. 
Hence, we define how much an event $e_k$ contributes to edge strength by means of the IWE \eqref{eq:IWE}, $^1 c_k$.

\textbf{Invariance}. 
Note that \eqref{eq:denoising} is invariant to monotonically increasing functions of $c_k$, %
as such functions preserve the order.
The same $\cEsig$ is obtained if one ranks and thresholds (at the corresponding level) values $\{c_k\}$, $\{\ln(c_k)\}$, $\{e^{c_k}\}$, etc. 
They are equivalent under the ranking rule \eqref{eq:denoising}. 

\textbf{Robustness to Various Edge Strengths}.
The Gaussian kernel in the IWE controls the sensitivity to edge strength.
Since the proposed method is based on both signal and noise IWEs,
increasing the size of the Gaussian kernel in the IWE
can emphasize contour edges over isolated points.
This preserves signal events better in regions with low IWE intensities,
for example at pixels with large depth values.

\subsection{Alternating Optimization}
\label{sec:method:iteration}
Classifying events into signal and noise requires knowledge of the true motion, 
and estimation of the true motion (e.g., using CMax) requires knowledge of the signal events because, by definition, noise events carry no information about motion. 
Hence, it is a circular dependency problem.

We approach its solution iteratively, as shown in \cref{fig:eyeCatcher}, by combining motion estimation (\cref{sec:method:cmax}) and S/N classification (\cref{sec:method:denoise}). 
Assuming some initialization, the current signal events $\cEsig$ are used to estimate the motion via CMax (1 step is enough);
then all events are warped to compute the IWE and the per-event scores $c_k$, from which the sets \eqref{eq:denoising} are recomputed, in preparation for the next iteration.
In practice, we initialize with a random split of $\cE$ into $\cEsig$ and $\cEnoi$,
update $\cEsig$ at every warp during the optimization iteration until the process converges (as flagged by the convergence of the motion parameters in CMax).

\textbf{Computational Complexity}. 
The computational complexity of one iteration of the proposed method is
$O(\numPixels + \numEvents \log\numEvents)$, which is slightly larger than the original CMax (i.e., $O(\numPixels + \numEvents)$) due to the search for the highest-ranked events during the optimization.

\section{Experiments}
\label{sec:experim}

We evaluate the proposed method on four datasets using various metrics (in \cref{sec:experim:datasets}).
In the following sections using common benchmarks,
we assess denoising (\cref{sec:experim:denoise}),
joint motion estimation (\cref{sec:experim:angVel}),
and the efficacy on intensity reconstruction (\cref{sec:experim:intensityRecon}).
We also conduct sensitivity analyses (\cref{sec:sensitivity}) and an ablation (\cref{sec:experim:dnn}).

\subsection{Datasets, Metrics, and Hyper-parameters}
\label{sec:experim:datasets}

\textbf{Datasets}.
\textbf{E-MLB} \cite{Ding23tom} is a large-scale, de-facto dataset to benchmark denoising.
It consists of $100$ sequences with 4 different brightness levels using neutral-density (ND) filters on a DAVIS346 camera ($346 \times 260$ px) during day and night.

Historically, the two sequences (\emph{hotel} and \emph{driving}) of the \textbf{DND21} dataset \cite{Guo22pami} have been widely used for denoising evaluation \cite{Guo22pami,Ding23tom,Jiang24eccv}.
Both sequences are recorded with a DAVIS346 camera \cite{Taverni18tcsii}.
They are regarded as ``signal'' data, as the sequences have been aggressively filtered offline.
Combined with other pure BA noise sequences, 
the dataset has event-wise noise annotations that are useful for evaluation.
We use the above sequences with different noise rates, $1,3,5,7,10$~Hz per pixel, following prior work \cite{Jiang24eccv}.

\sisetup{round-mode=places,round-precision=3}
\begin{table}[t]
\centering
\adjustbox{max width=\linewidth}{%
\setlength{\tabcolsep}{3pt}
\renewcommand{\arraystretch}{1.2} %
\begin{tabular}{ll*{9}{S[table-format=2.3]}}
\toprule
& & \multicolumn{4}{c}{E-MLB (Day)}
& \multicolumn{4}{c}{E-MLB (Night)}
& \text{DND21}
 \\
 \cmidrule(l{1mm}r{1mm}){3-6}
 \cmidrule(l{1mm}r{1mm}){7-10}
& & \text{ND1} & \text{ND4} & \text{ND16} & \text{ND64} 
& \text{ND1} & \text{ND4} & \text{ND16} & \text{ND64} 
\\
\midrule
\multirow{10}{*}{\rotatebox{90}{\makecell{Model-based}}} 
& Raw & 0.821 & 0.824 & 0.815 & 0.786 & 0.89 & 0.824 & 0.786 & 0.768 & 0.869 \\
& BAF \cite{Delbruck08issle} & 0.861 & 0.869 & 0.876 & 0.89 & 0.946 & \unumr{1.3}{0.973} & \bnum{0.992} & \bnum{0.942} & 0.92 \\
& TS \cite{Lagorce17pami} & 0.877 & 0.887 & 0.87 & 0.837 & 1.033 & 0.944 & 0.886 & 0.797 & 0.985 \\
& KNoise \cite{Khodamoradi18tetc} & 0.846 & 0.837 & 0.83 & 0.807 & 0.954 & 0.956 & 0.871 & 0.817 & 0.887 \\
& EvFlow \cite{Wang19cvpr} & 0.848 & 0.878 & 0.868 & 0.833 & 0.969 & \bnum{0.983} & \unumr{1.3}{0.889} & 0.797 & \bnum{1.006} \\
& IETS \cite{Baldwin19iciar} & 0.772 & 0.785 & 0.777 & 0.753 & 0.950 & 0.823 & 0.804 & 0.711 & 0.900 \\
& Ynoise \cite{Feng20ap} & 0.866 & 0.863 & 0.857 & 0.821 & 1.009 & 0.943 & 0.875 & 0.792 & 0.966 \\
& GEF \cite{Duan21pami} & \bnum{1.051} & \unumr{1.3}{0.938} & \unumr{1.3}{0.935} & \unumr{1.3}{0.927} & \unumr{1.3}{1.027} & 0.955 & 0.946 & \unumr{1.3}{0.935} & 0.932 \\
& DWF \cite{Guo22pami} & 0.878 & 0.876 & 0.866 & 0.865 & 0.923 & 0.962 & 0.988 & 0.932 & 0.905 \\
& \textbf{Ours} & \unumr{1.3}{0.9382543842929747} & \bnum{0.9584643352770248} & \bnum{0.9858964501229918} & \bnum{0.9500818802404596} & \bnum{1.036831015641091} & 0.961426679598107 & 0.9445279530225384 & 0.932362037844126 &  \unumr{1.3}{0.9920691211} \\
\cmidrule(l{1mm}r{1mm}){1-11}

\multirow{4}{*}{\rotatebox{90}{\makecell{Learning}}} 
& EDnCNN \cite{Baldwin20cvpr} & 0.887 & 0.908 & 0.903 & 0.912 & 1.001 & \bnum{1.024} & \bnum{1.079} & \unumr{1.3}{1.086} & 0.977 \\
& EventZoom \cite{Duan21cvpr} & \bnum{0.996} & \bnum{0.988} & \bnum{0.996} & \bnum{0.97} & \bnum{1.055} & 1.007 & 1.01 & 0.988 & \bnum{1.059} \\
& MLPF \cite{Guo22pami} & 0.851 & 0.855 & 0.846 & 0.84 & 0.926 & 0.928 & 0.91 & 0.906 & 0.944 \\
& EDformer \cite{Jiang24eccv} & \unumr{1.3}{0.952} & \unumr{1.3}{0.955} & \unumr{1.3}{0.956} & \unumr{1.3}{0.942} & \unumr{1.3}{1.048} & \unumr{1.3}{1.019} & \unumr{1.3}{1.076} & \bnum{1.099} & \unumr{1.3}{1.041} \\
\bottomrule
\end{tabular}
}
\caption{\label{tab:mesr}Mean ESR (MESR$\uparrow$) results among denoising methods on the event denoising datasets E-MLB \cite{Ding23tom} and DND21 \cite{Guo22pami}. 
In each category, the best is in bold and the second best is underlined.
Methods are chronologically sorted within each category.
}
\end{table}

\sisetup{round-mode=places,round-precision=3}
\begin{table}[t]
\centering
\adjustbox{max width=\linewidth}{%
\setlength{\tabcolsep}{4pt}
\begin{tabular}{ll*{10}{S[table-format=2.3]}}
\toprule
 &  & \multicolumn{2}{c}{1Hz} 
 & \multicolumn{2}{c}{5Hz} 
 & \multicolumn{2}{c}{10Hz}\\
\cmidrule(lr){3-4}\cmidrule(lr){5-6}\cmidrule(lr){7-8}
 &  & \text{\emph{hotel}} & \text{\emph{driving}}  
 & \text{\emph{hotel}}  & \text{\emph{driving}}  
 & \text{\emph{hotel}}  & \text{\emph{driving}} \\
\midrule 
\multirow{6}{*}{\rotatebox{90}{\makecell{Model-based}}} 
 & BAF \cite{Delbruck08issle}  & 0.9535  & 0.8479  & 0.8916  & 0.7930  & 0.8366  & 0.7479 \\
 & TS \cite{Lagorce17pami}  & \unumr{1.3}{0.9716}  & \unumr{1.3}{0.9307}  & \unumr{1.3}{0.9606}  & \bnum{0.9270}  & \bnum{0.9620}\  & \bnum{0.9202} \\
 & KNoise \cite{Khodamoradi18tetc}  & 0.6773  & 0.6296  & 0.6703  & 0.6235  & 0.6413  & 0.6142 \\
 & Ynoise \cite{Feng20ap}  & 0.9690  & \bnum{0.9409}  & 0.9234  & \unumr{1.3}{0.9093}  & 0.8987  & \unumr{1.3}{0.8800} \\
 & DWF \cite{Guo22pami}  & 0.9268  & 0.7409  & 0.8620  & 0.6901  & 0.7958  & 0.6563\\
 & \textbf{Ours}  & \bnum{1.0138021943171762}  & 0.8821196931336948  & \bnum{0.9625158156194444}  & 0.8553984846381162  & \unumr{1.3}{0.9605647249996442}  & 0.836273555102236 \\
\cmidrule(lr){1-8}

\multirow{3}{*}{\rotatebox{90}{\makecell{Learning}}} 
 & EDnCNN \cite{Baldwin20cvpr}  & 0.9573  & 0.8873  & 0.9365  & 0.8748  & 0.9006  & 0.874 \\
 & MLPF \cite{Guo22pami}  & \unumr{1.3}{0.9704}  & \unumr{1.3}{0.8887}  & \unumr{1.3}{0.9704}  & \unumr{1.3}{0.8845}  & \unumr{1.3}{0.9634}  & \unumr{1.3}{0.8761} \\
 & EDformer \cite{Jiang24eccv}  & \bnum{0.9928}  & \bnum{0.9541}  & \bnum{0.9845}  & \bnum{0.9424}  & \bnum{0.9699}  & \bnum{0.9264} \\
\bottomrule
\end{tabular}
}
\caption{\label{tab:dnd21_roc}The AUC$\uparrow$ of ROC on the two DND21 sequence at different noise rates. 
We follow \cite{Jiang24eccv} for baseline methods selection.}
\end{table}

\begin{figure}[t]
  \centering
  {{\includegraphics[clip,trim={0.6cm 0.7cm 0.7cm 0.7cm},width=\linewidth]{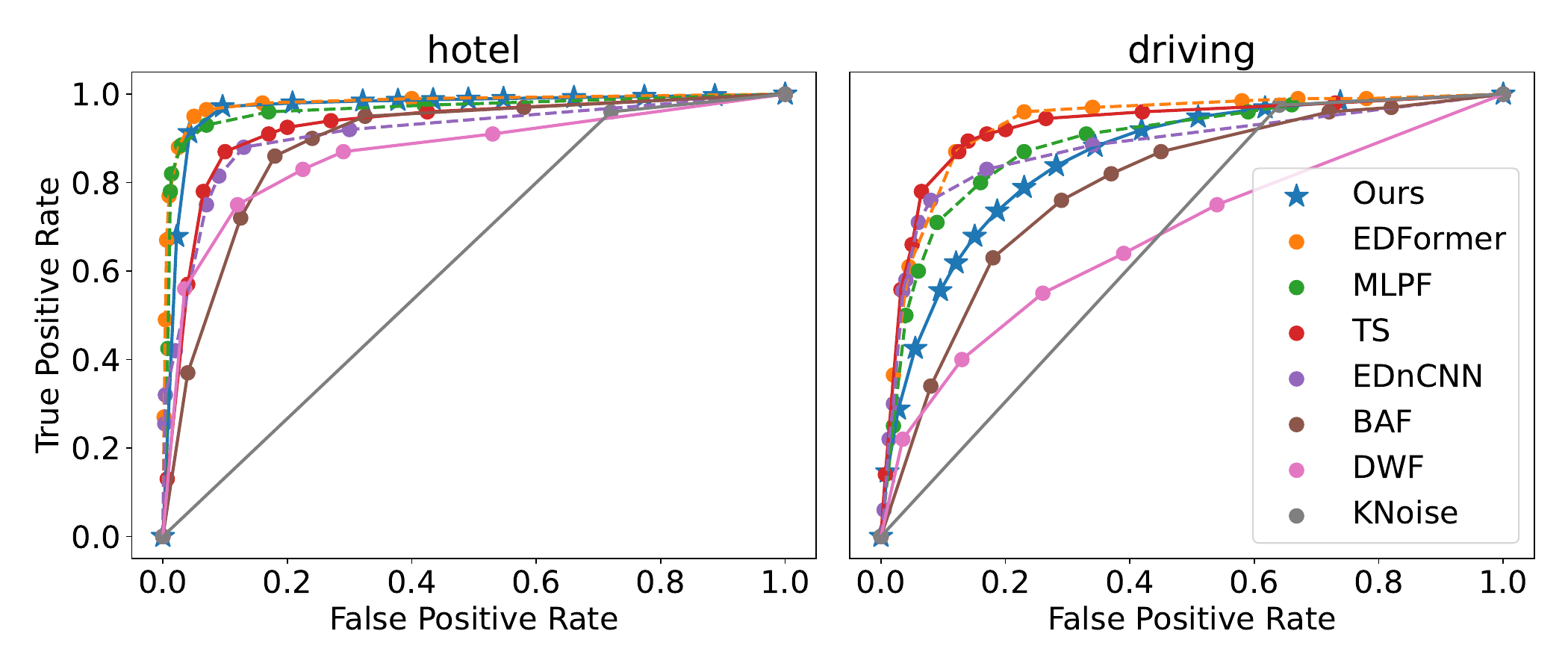}}}
\caption{ROC curves on the DND 21 dataset sequences. Dash lines represent learning-based methods (see also \cref{tab:dnd21_roc}).
}
\label{fig:result:dnd21roc}
\vspace{-2ex}
\end{figure}

\begin{figure*}[t]
	\centering
		{\includegraphics[clip,trim={0 8.7cm 1cm 0cm},width=\linewidth]{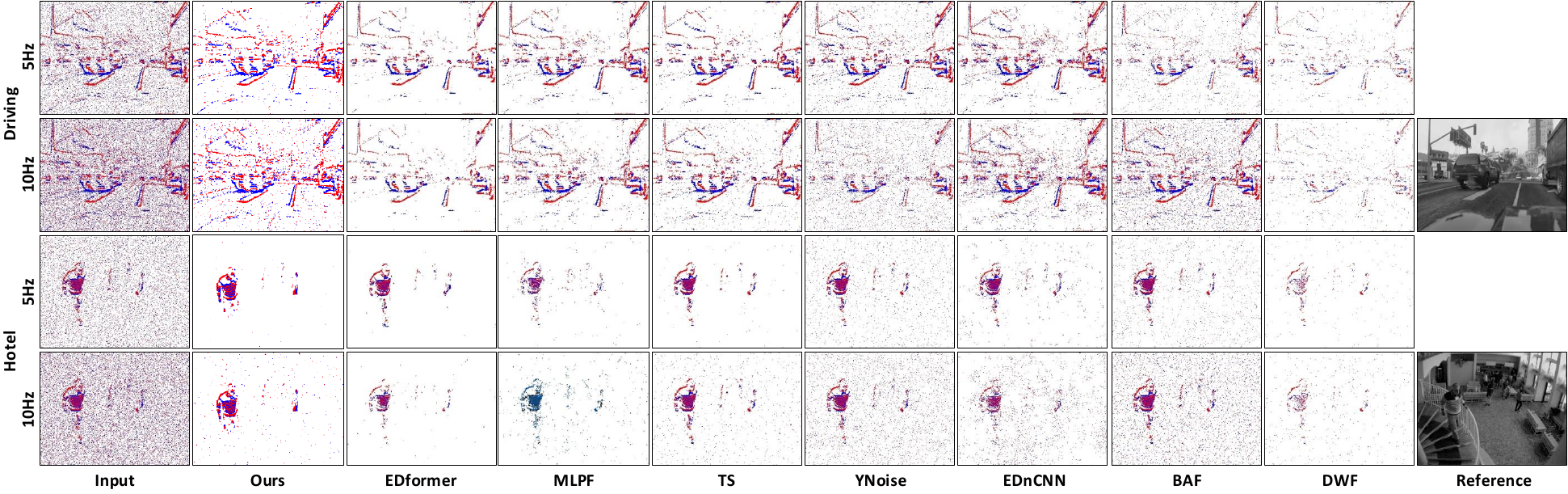}}
    \vspace{-3ex}
	\caption{Comparison of denoising methods on the two DND21 sequences %
    at two different noise levels. 
    }
    \label{fig:result:dnd21Comparison}
    \vspace{-1ex}
\end{figure*}

The \textbf{ECD} dataset \cite{Mueggler17ijrr} is a standard dataset for various tasks including camera ego-motion estimation \cite{Gallego17ral,Zhu17cvpr,Rosinol18ral,Mueggler18tro,Zhou20tro,Shiba22sensors}.
Using a DAVIS240C camera ($240 \times 180$ px \cite{Brandli14ssc}),
each sequence provides events, frames, calibration information, IMU data,
and ground truth (GT) camera poses (at $200$~Hz).
We use \emph{boxes\_rotation} and \emph{dynamic\_rotation} sequences to evaluate rotational motion estimation accuracy.

Finally, we also use sequences from \textbf{MVSEC} \cite{Zhu18ral} and \emph{zurich\_city\_12a} sequence from the \textbf{DSEC} dataset \cite{Gehrig21ral},
which are typically used for optical flow estimation.
We use MVSEC sequences for quantitative evaluation and the DSEC sequence for qualitative evaluation, since it is captured during the night and hence includes lots of noise.

\textbf{Metrics}. 
The metrics used for denoising assessment are: 
receiver operating characteristic (ROC) curve, area under the curve (AUC) and Mean Event Structural Ratio (MESR).
Given signal and noise GT labels, the ROC shows the True-Positive Rate and False-Positive Rate over several S/N ratios.
The AUC of the ROC becomes closer to $1$, the better the denoiser works.
The MESR is a denoising metric that does not require GT labels \cite{Ding23tom}.
We follow previous work to fix the number of signal events to 30000 \cite{Jiang24eccv,Ding23tom}, as well as  $M = 20000$ \cite[Eq.10]{Ding23tom},
to discount the difference in the number of signal events from various denoising methods.
The higher, the better.
Furthermore, we use the RMS error for angular velocity estimation and the EPE for optical flow estimation, following previous works \cite{Nunes21pami,Gu21iccv,Guo24tro,Shiba24pami},
as well as the FWL metric \cite{Stoffregen20eccv}, which is the relative variance (sharpness) of the IWE with respect to that of the identity warp.
Further details about the experimental settings (e.g., hyper-parameters) are in the supplementary.

\subsection{Denoising Results}
\label{sec:experim:denoise}

We evaluate denoising accuracy using AUC on DND21 data, and using MESR on E-MLB and DND21 data,
using the optical flow estimation.

\textbf{Denoising Performance on E-MLB}.
\Cref{tab:mesr} shows the benchmark results on the E-MLB dataset using the MESR metric.  
We categorize prior work into model-based methods and learning-based methods for convenience.
Compared with other model-based techniques,
our method ranks first or second,
showcasing the efficacy of the proposed method on denoising. %
Even compared with learning-based ones, our method achieves competitive scores (e.g., second in E-MLB (Day) ND4--ND64 conditions, etc).
Notice that, however, learning-based approaches utilize additional information (GT labels) during training, while our method does not.
Nonetheless, the results demonstrate the efficacy of our approach,
which contributes to the theoretical foundation of event-denoising methodologies.
We discuss further details of the evaluation metrics (ESR) in \cref{sec:limitations}.

Qualitative results are shown in \cref{fig:result:denoiseQuality}.
The denoised outputs (``Signal'') are reasonable, as dark (underexposed) regions of the scene become cleaner,
and the IWEs using the estimated flow provide sharp edges of the original scenes.
We compare with state-of-the-art methods in detail next.

\def\figWidth{0.3\linewidth}
\def\textWidth{1em}
\begin{figure}[t]
	\centering
    {\scriptsize
    \setlength{\tabcolsep}{1pt}
        \begin{tabular}{
	>{\centering\arraybackslash}m{\textWidth}
	>{\centering\arraybackslash}m{\figWidth} 
	>{\centering\arraybackslash}m{\figWidth}
	>{\centering\arraybackslash}m{\figWidth}}
		\\

            &\gframe{\includegraphics[width=\linewidth]{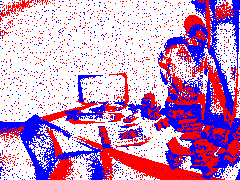}}
            &\gframe{\includegraphics[width=\linewidth]{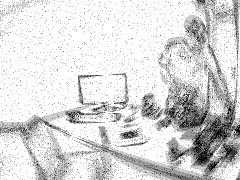}}
            &\gframe{\includegraphics[width=\linewidth]{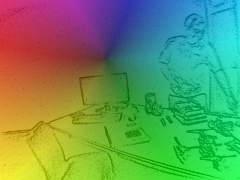}} \\
            & Raw Events & IWE(0) & GT rotation \\[0.5ex]

            \rotatebox{90}{\makecell{IWE of signal}}
		&\gframe{\includegraphics[width=\linewidth]{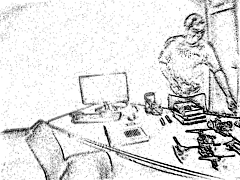}}
		&\gframe{\includegraphics[width=\linewidth]{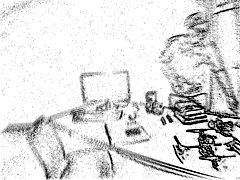}}
		&\gframe{\includegraphics[width=\linewidth]{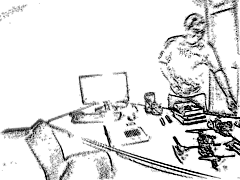}}
		\\

            \rotatebox{90}{\makecell{Estimated rotation}}
		&\gframe{\includegraphics[width=\linewidth]{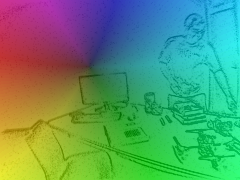}}
		&\gframe{\includegraphics[width=\linewidth]{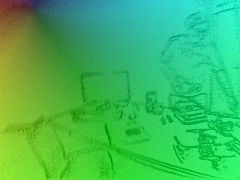}}
		&\gframe{\includegraphics[width=\linewidth]{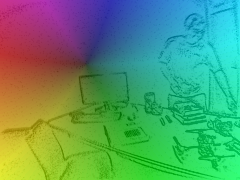}}
		\\

            \rotatebox{90}{\makecell{Signal events}}
		&\gframe{\includegraphics[width=\linewidth]{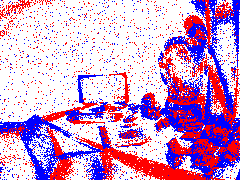}}
		&\gframe{\includegraphics[width=\linewidth]{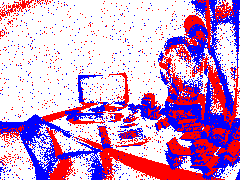}}
		&\gframe{\includegraphics[width=\linewidth]{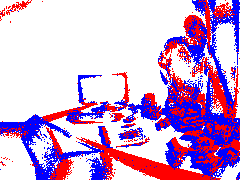}}
		\\
  
            \rotatebox{90}{\makecell{Noise events}}  
		&\gframe{\includegraphics[width=\linewidth]{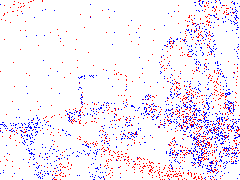}}
		&\gframe{\includegraphics[width=\linewidth]{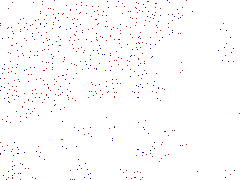}}
		&\gframe{\includegraphics[width=\linewidth]{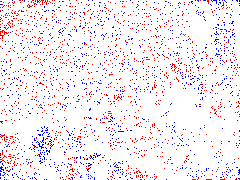}}
		\\  

		& (a) Random sampling
		& (b) BA filter (BAF) \cite{Delbruck08issle}
		& (c) Ours
		\\
	\end{tabular}
	}
	\caption{Simultaneous estimation of rotational motion and denoising on the \emph{dynamic\_rotation} sequence from ECD dataset \cite{Mueggler17ijrr}.
    Our method produces sharp IWEs (\nth{2} row) as well as reasonable S/N classification (rows 4 and 5), 
    keeping 90\% of data ($\snratio = 0.9$). 
    }
\label{fig:result:angvelComp}
\vspace{-2ex}
\end{figure}

\textbf{Denoising Accuracy on DND21}.
\Cref{tab:dnd21_roc} shows the result of AUC for the ROC result on the DND21 dataset.
Our method achieves consistently high AUC compared with existing model-based methods.
Note that the DND21 dataset aggressively removes events to generate ``pure-signal'' sequences \cite[Sec.5]{Guo22pami}.
Due to this step, the labeled signal events may be fewer than the actual ones corresponding to the true motion.
Hence, the DND21 sequences could result in a slight degradation in the False Positive score for motion-based (unsupervised) denoising methods like ours,
which may result in lower AUC values.
Nonetheless, our method provides competitive AUC scores among other state-of-the-art.
The ROC curves are shown in \cref{fig:result:dnd21roc}.

\begin{figure}[t]
  \centering
  {{\includegraphics[clip,trim={0cm 2.5cm 12.5cm 0cm},width=\linewidth]{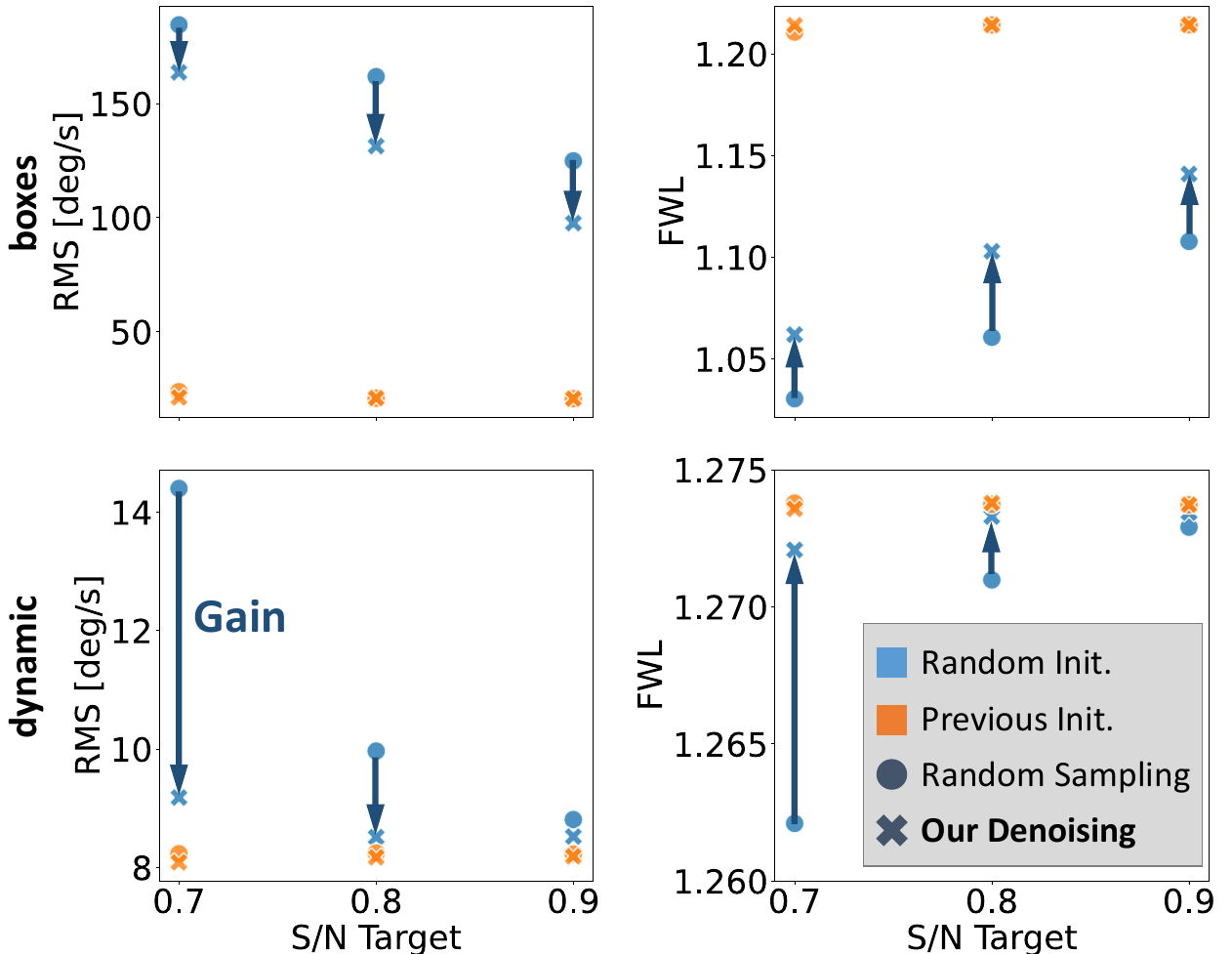}}}
	\caption{Results on the ego-motion estimation application (top: \emph{boxes}, bottom: \emph{dynamic}) \cite{Mueggler17ijrr}. 
    Our method relaxes the dependency on the initial value for CMax-based rotational motion estimation (indicated as ``Gain'').}
    \label{fig:result:angvel_quantitative}
\vspace{-2ex}
\end{figure}

\def\figWidth{0.14\linewidth}
\begin{figure*}[t]
	\centering
    {\footnotesize
    \setlength{\tabcolsep}{1pt}
	\begin{tabular}{
	>{\centering\arraybackslash}m{\figWidth} 
	>{\centering\arraybackslash}m{\figWidth} 
	>{\centering\arraybackslash}m{\figWidth} 
	>{\centering\arraybackslash}m{\figWidth} 
	>{\centering\arraybackslash}m{\figWidth} 
	>{\centering\arraybackslash}m{\figWidth}}

        \gframe{\includegraphics[width=\linewidth]{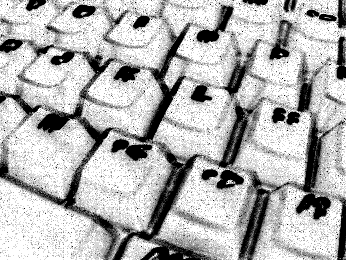}}
        & \gframe{\includegraphics[width=\linewidth]{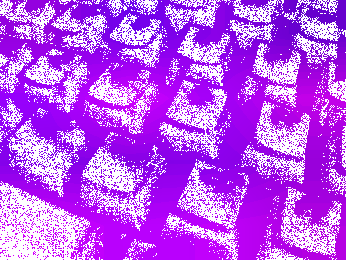}}
        & \gframe{\includegraphics[width=\linewidth]{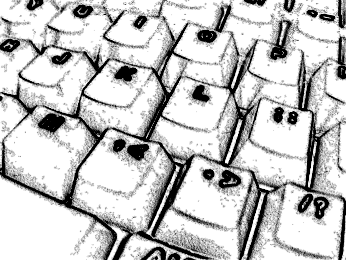}}
		&\gframe{\includegraphics[width=\linewidth]{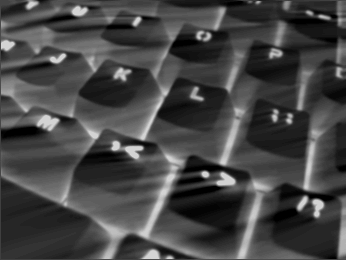}}
		&\gframe{\includegraphics[width=\linewidth]{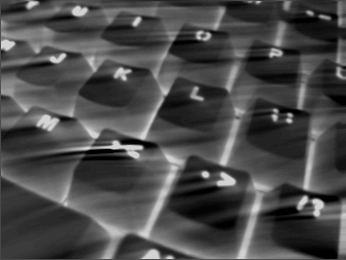}}	%
		&\gframe{\includegraphics[width=\linewidth]{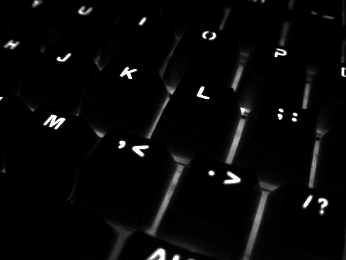}}
        \\

        \gframe{\includegraphics[width=\linewidth]{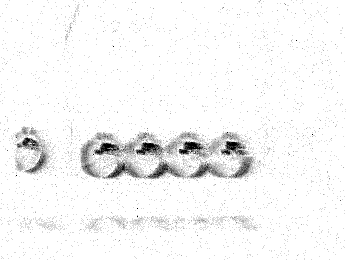}}
        & \gframe{\includegraphics[width=\linewidth]{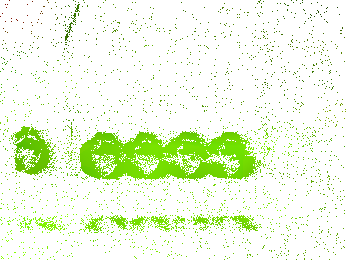}}
        & \gframe{\includegraphics[width=\linewidth]{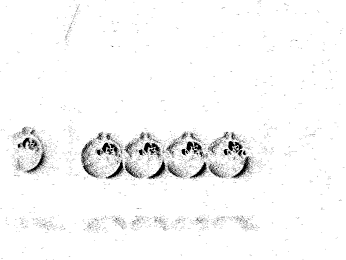}}
		&\gframe{\includegraphics[width=\linewidth]{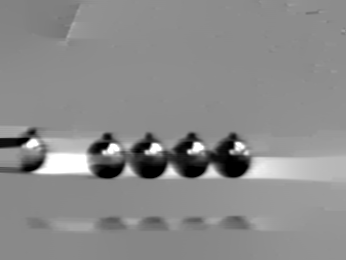}}
		&\gframe{\includegraphics[width=\linewidth]{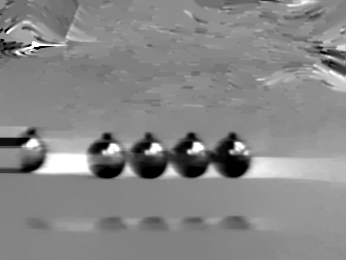}}
		&\gframe{\includegraphics[width=\linewidth]{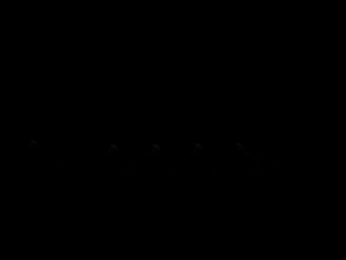}}
        \\

        \gframe{\includegraphics[width=\linewidth]{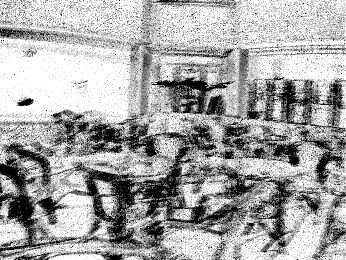}}
        & \gframe{\includegraphics[width=\linewidth]{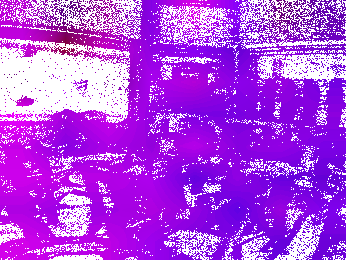}}
        & \gframe{\includegraphics[width=\linewidth]{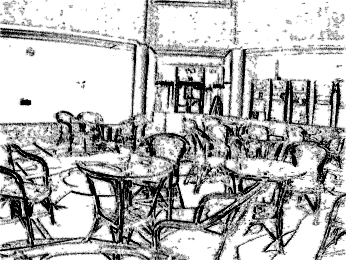}}
		&\gframe{\includegraphics[width=\linewidth]{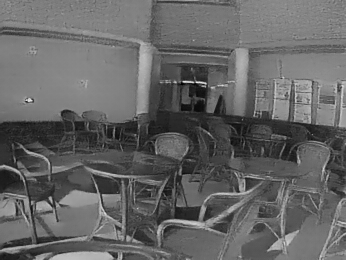}}
		&\gframe{\includegraphics[width=\linewidth]{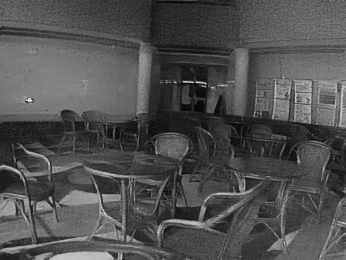}}	%
		&\gframe{\includegraphics[width=\linewidth]{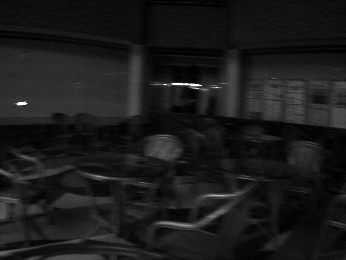}}
        \\

        \gframe{\includegraphics[width=\linewidth]{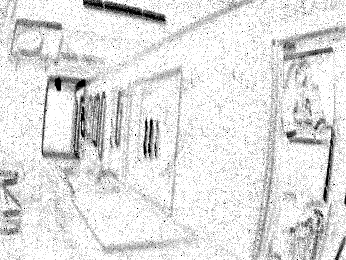}}
        & \gframe{\includegraphics[width=\linewidth]{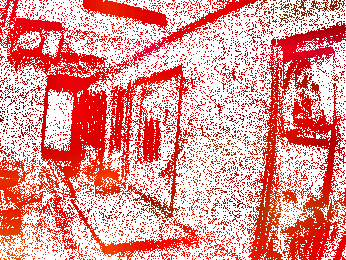}}
        & \gframe{\includegraphics[width=\linewidth]{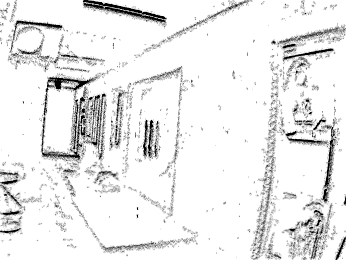}}
		&\gframe{\includegraphics[width=\linewidth]{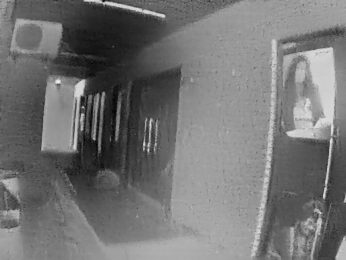}}
		&\gframe{\includegraphics[width=\linewidth]{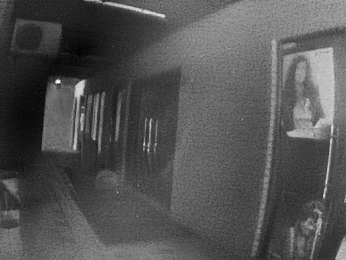}}	%
		&\gframe{\includegraphics[width=\linewidth]{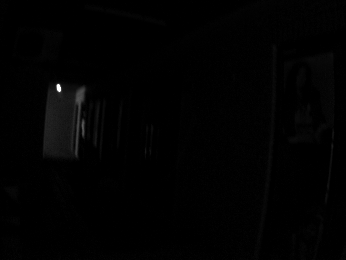}}
        \\

		(a) Input events
		& (b) Estim. flow
        & (c) Signal events
		& \makecell[t]{(d) Reconstr.\\ from (c) signal}
		& \makecell[t]{(e) Reconstr.\\ from (a) input}
		& (f) Reference frame
		\\
	\end{tabular}
	}
    \vspace{-1ex}
	\caption{\emph{Image intensity reconstruction} on E-MLB data \cite{Ding23tom}. 
    The first two rows use the reconstruction method EVILIP \cite{Zhang22pami} and the last two rows use E2VID \cite{Rebecq19pami}.
    Reconstruction on denoised events produces higher quality images (i.e., fewer artifacts) than on raw event data.}
\label{fig:result:endRecon}
\vspace{-2ex}
\end{figure*}

\sisetup{round-mode=places,round-precision=3}
\begin{table}[t]
\centering
\adjustbox{max width=.9\linewidth}{%
\setlength{\tabcolsep}{4pt}
\begin{tabular}{ll*{4}{S[table-format=2.3]}}
\toprule
& & \multicolumn{2}{c}{\emph{dynamic\_rot}} & \multicolumn{2}{c}{\emph{boxes\_rot}}
 \\
 \cmidrule(l{1mm}r{1mm}){3-4}
 \cmidrule(l{1mm}r{1mm}){5-6}
& &\text{RMS $\downarrow$} & \text{FWL $\uparrow$} &\text{RMS $\downarrow$} & \text{FWL $\uparrow$}
\\
\midrule
\multirow{8}{*}{\begin{turn}{90}
$5$~Hz Noise
\end{turn}} 
& CMax \cite{Gallego18cvpr} & 20.00070968 & 1.258796097 & 124.641377 & 1.128529593 \\ 
& -- w/ Init. & 8.274876245 & 1.273697045 & 20.65924033 & 1.214159459 \\
& Downsampling (Best) & 8.80770395 & 1.272907822 & 124.8686397 & 1.107758395 \\ 
& -- w/ Init. & \unumr{1.3}{8.226219444} & \unumr{1.3}{1.273725307} & 20.61861586 & \unumr{1.3}{1.214188455} \\
& Ours (Best) & 8.521574073 & 1.273446296 & 97.58457484 & 1.14088066 \\ 
& -- w/ Init. & \bnum{8.170} & \bnum{1.273725299} & \unumr{2.3}{20.60433462} & \bnum{1.214179018} \\
& BAF  & 19.67489305 & 1.25977249 & 125.0281451 & 1.127308902 \\ 
& -- w/ Init. & 8.253078202 & 1.27363699 & \bnum{20.5504366} & 1.21350548 \\ 
\bottomrule
\end{tabular}
}
\caption{\label{tab:angvelShort}Angular velocity estimation on ECD dataset \cite{Mueggler17ijrr}.}
\vspace{-2ex}
\end{table}

Qualitative comparisons among several denoising %
methods is given in \cref{fig:result:dnd21Comparison}.
From the results, our method achieves
a good trade-off between under-denoising (preserving not only scene-edge details but some noise)
and over-denoising (removing not only most of the noise but some edge details).
This can be confirmed, for example, in the upper-right edges in the driving example, and the person's arm in the hotel example.
Essentially, all denoising methods have some parameters to control the amount of denoising;
our proposed method has the signal ratio, which is intuitive; 
other optimization methods have some parameters about spatio-temporal kernel size,
and learning-based approaches use the probability threshold for the network output. %
While we fix it for each dataset and condition for fair comparison (we do not tune it for each sequence in the results),
this is an essential challenge for denoising algorithms in general, regardless of their underlying character (learning-based or model-based). 
Hence, we believe it is important for research to discuss denoising effectiveness in combination with downstream tasks, as shown in \cref{sec:experim:angVel,sec:experim:intensityRecon}.

\subsection{Joint Motion Estimation Results}
\label{sec:experim:angVel}

\textbf{Ego-Motion on ECD.}
One of the merits of the proposed joint estimation method is that it can improve motion estimation results. 
Here, the results of ego-motion (e.g., angular velocity) estimation are displayed in \cref{fig:result:angvelComp}.
We used the classical ECD dataset and injected BA noise from the DND21 recordings with different noise rates.
As no prior work simultaneously estimates motion and noise, we compare against sequential filters:
random sampling (of the input data with a uniform distribution)
and BAF \cite{Delbruck08issle} that is the most widely-used denoising method to date.
As shown in the IWEs, our method produces the sharpest edges (cf.~(b) and (c)) 
and the most reasonable signal-noise classification among others (cf.~(a), (b), and (c)).
Note that the original ECD recordings have noise events besides the ones that are injected using DND21 recordings,
which makes the calculation of ROC (similar to \cref{sec:method:denoise}) intricate.
Nonetheless, we report ROC results in the supplementary.

The quantitative evaluation on angular velocity estimation is reported in \cref{fig:result:angvel_quantitative},
as well as the summary of the supplementary detailed table in \cref{tab:angvelShort}.
As shown in the first row (``CMax''),
the original CMax degrades due to the noise occurrences.
Our method achieves the best results in terms of RMS and FWL,
compared with the random sampling method as well as the original CMax (no denoising).
Also, using better initialization (the result of the previous slice for sequential estimation) always improves the results (``w/ Init.''):
the estimation accuracy relies on both the initialization and the denoising strategies.
Hence, the result demonstrates that the proposed method relaxes the dependency on the initialization, 
thus improving the robustness of motion estimation.

\gblue{\textbf{Optical Flow on MVSEC.}
The results on more complex motion estimation (e.g., optical flow) are shown in \cref{tab:flowMvsec}, using the MVSEC dataset \cite{Zhu18ral},
with BA noise from the DND21 dataset.
Due to the $5$Hz BA noise, the baseline method MultiCM \cite{Shiba22eccv} degrades. 
Our denoising method improves the flow accuracy, showcasing the efficacy of simultaneous noise estimation.
A challenge of evaluating denoising in these real-world datasets (e.g., ECD, MVSEC) is the existing noise in the original sequences.
Hence, we only evaluate motion estimation metrics, not ROC or AUC.}

\sisetup{round-mode=places,round-precision=3}
\begin{table}[t]
\centering
\adjustbox{max width=.7\linewidth}{%
\setlength{\tabcolsep}{4pt}
\begin{tabular}{l*{4}{S[table-format=2.3]}}
\toprule
 & \multicolumn{2}{c}{\emph{indoor1}} & \multicolumn{2}{c}{\emph{indoor2}}
 \\
 \cmidrule(l{1mm}r{1mm}){2-3}
 \cmidrule(l{1mm}r{1mm}){4-5}
 &\text{EPE $\downarrow$} & \text{\%Out $\downarrow$} &\text{EPE $\downarrow$} & \text{\%Out $\downarrow$}
\\
\midrule
MultiCM \cite{Shiba22eccv} & 2.675668117 & 32.09846481 & 2.746023648 & 33.47406857 \\ 
Ours & \bnum{2.516871532} & \bnum{29.47534065} & \bnum{2.695166225} & \bnum{33.04673993} \\ 
\midrule
 & \multicolumn{2}{c}{\emph{indoor3}} & \multicolumn{2}{c}{\emph{outdoor1}}
 \\
 \cmidrule(l{1mm}r{1mm}){2-3}
 \cmidrule(l{1mm}r{1mm}){4-5}
 &\text{EPE $\downarrow$} & \text{\%Out $\downarrow$} &\text{EPE $\downarrow$} & \text{\%Out $\downarrow$}
\\
\midrule
MultiCM \cite{Shiba22eccv} & 2.73172967 & 33.21893744 & 1.960064541 & 18.15950336 \\ 
Ours & \bnum{2.653875277} & \bnum{32.24148888} & \bnum{1.86462922} & \bnum{17.55410014} \\ 
\bottomrule
\end{tabular}
}
\caption{\label{tab:flowMvsec} \emph{Optical flow estimation results} on MVSEC \cite{Zhu18ral} ($dt=4$ frames, i.e., 89 ms) with BA noise at $5$~Hz.}
\vspace{-2ex}
\end{table}

\subsection{Application to Intensity Reconstruction}
\label{sec:experim:intensityRecon}

In real-world use cases,
denoising works as preprocessing steps for other tasks (e.g., deblur, 3D reconstruction, motion estimation, etc.),
hence it is important for denoising methods to demonstrate applicability.
Although the proposed approach intrinsically improves motion estimation (\cref{sec:experim:angVel}), 
we further evaluate the denoising outcome
via intensity reconstruction, a well-known task in event-based vision, using the E-MLB dataset \cite{Ding23tom}.
\Cref{fig:result:endRecon} collects the results.
We use two state-of-the-art approaches for reconstruction:
E2VID \cite{Rebecq19pami}, which does not rely on motion, 
and EVILIP \cite{Zhang22pami}, which utilizes optical flow as prior.

Our method provides flow (\cref{fig:result:endRecon},
column (b)) and signal events (sharp IWEs in column (c)).
Comparing the reconstruction with our event denoising method (column (d)) and the reconstruction from the raw event data (column (e)),
the images after event denoising have fewer artifacts than those without denoising, for both reconstruction methods tested.
We observe that E2VID is more robust against noise than EVILIP, possibly due to noise events in its training (simulation) data.
Due to the ND filters, the quality of the reference frames (last column) is limited, and hence, we discuss the results qualitatively. 
Nevertheless comparing the reconstructed images to the reference frames, evidences the HDR advantages of event cameras over frame-based ones.

\def\figWidth{0.3\linewidth}
\def\textWidth{1em}
\begin{figure}[t]
	\centering
    {\scriptsize
    \setlength{\tabcolsep}{1pt}

        \begin{tabular}{
	>{\centering\arraybackslash}m{\textWidth}
	>{\centering\arraybackslash}m{\figWidth} 
	>{\centering\arraybackslash}m{\figWidth}
	>{\centering\arraybackslash}m{\figWidth}}
		\\

            \rotatebox{90}{\makecell{$\snratio = 0.9$}}
            &\gframe{\includegraphics[width=\linewidth]{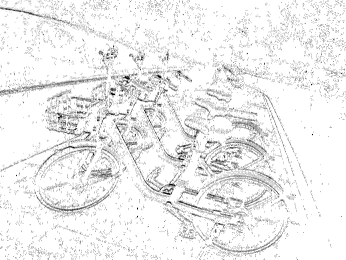}}
            &\gframe{\includegraphics[width=\linewidth]{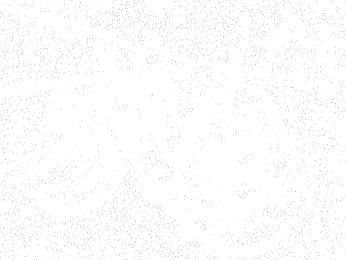}}
            &\gframe{\includegraphics[width=\linewidth]{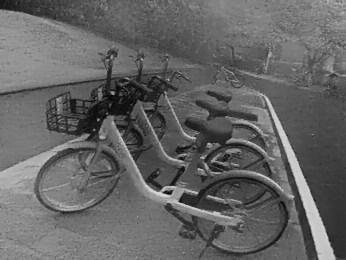}} \\

            \rotatebox{90}{\makecell{$\snratio = 0.7$}}  
            &\gframe{\includegraphics[width=\linewidth]{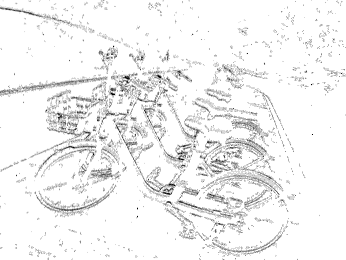}}
            &\gframe{\includegraphics[width=\linewidth]{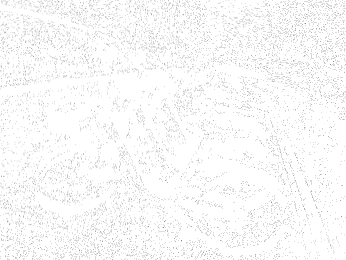}}
            &\gframe{\includegraphics[width=\linewidth]{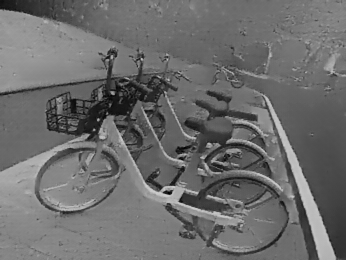}} \\

            \rotatebox{90}{\makecell{$\snratio = 0.5$}}            
            &\gframe{\includegraphics[width=\linewidth]{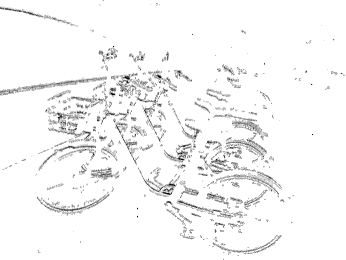}}
            &\gframe{\includegraphics[width=\linewidth]{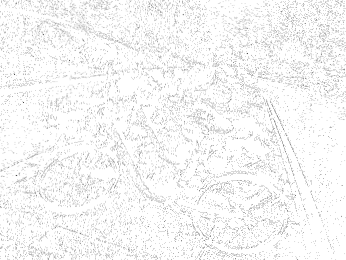}}
            &\gframe{\includegraphics[width=\linewidth]{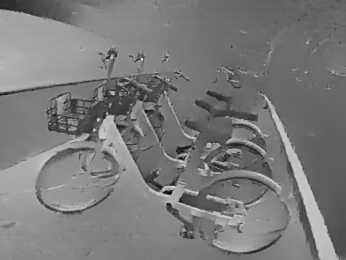}} \\
            
            & (a) Signal
            & (b) Noise
            & (c) Reconstruction
		\\[-1ex]
	\end{tabular}
	}
	\caption{Sensitivity analysis: effect of different target ratio parameters $\snratio$ on image reconstruction using EVILIP \cite{Zhang22pami}.
    }
\label{fig:sensitivity:recon}
\end{figure}

\subsection{Sensitivity}
\label{sec:sensitivity}

\textbf{Effect of Noise Ratio}. As \cref{sec:experim:denoise} already showed the sensitivity for denoising accuracy,
let us further discuss the sensitivity using image reconstruction.
\Cref{fig:sensitivity:recon} presents the result of the \emph{Bicycle} sequence with ND64 filter \cite{Ding23tom}.
While the GT image is highly underexposed and not informative (as those on the last column of \cref{fig:result:endRecon}),
the reconstructed images clearly reflect the choice of the signal target ratio $\snratio$.
As more events are removed, ($i$) the reconstructed image becomes more homogeneous but preserves strong edges,
and ($ii$) the removed events (2nd col.) contain more scene edges (i.e., signal).
We quantitatively analyze the former in the supplementary using non-reference image quality metrics.

\textbf{Effect of the Kernel Size}.
\Cref{fig:rebuttal:kernel} shows the sensitivity analysis for different kernel sizes.
By increasing the size of the Gaussian kernel in the IWE, our method can emphasize contour edges over isolated points, thus better preserving signal events in regions with low IWE intensities (center of the image, corresponding to far away scene points).

\def\figWidth{0.3\linewidth}
\begin{figure}[t]
	\centering
    {\scriptsize
    \setlength{\tabcolsep}{2pt}
	\begin{tabular}{
	>{\centering\arraybackslash}m{\figWidth} 
	>{\centering\arraybackslash}m{\figWidth} 
	>{\centering\arraybackslash}m{\figWidth}}

            \gframe{\includegraphics[width=\linewidth]{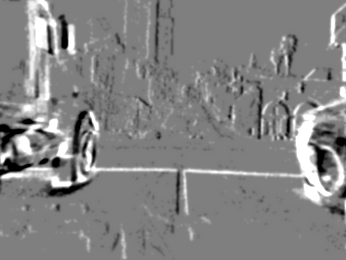}}
            & \gframe{\includegraphics[width=\linewidth]{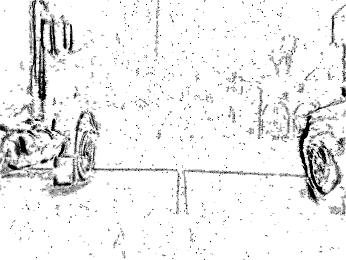}}
		&\gframe{\includegraphics[width=\linewidth]{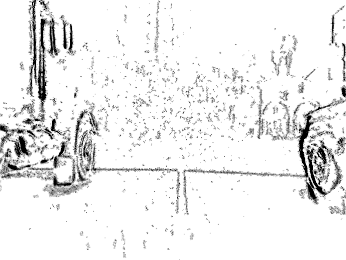}} \\

		(a) Ref. events
        & (b) $k = 1$ px
		& (c) $k = 3$ px \\[-2ex]
	\end{tabular}
	}
	\caption{\label{fig:rebuttal:kernel}Results of different Gaussian kernel sizes $k$ for sensitivity on different depth scenes. 
    \vspace{-2ex}}
\end{figure}

\subsection{Ablation of Motion Estimator}
\label{sec:experim:dnn}

The proposed approach can be combined with other estimation methods,
such as deep-neural--network (DNN) flow estimators.
To this end, we show an example of E-RAFT \cite{Gehrig21threedv}.
The model-based method alone \cite{Shiba22eccv} is known to output a degraded flow in night sequences with lots of noise, such as \emph{zurich\_city\_12a} \cite{Gehrig21ral}.
As shown in \cref{fig:result:appEraft}, our method can identify and discard noisy events.
E-RAFT provides sharper and more natural edges in IWEs after denoising compared to MultiCM \cite{Shiba22eccv} (thanks to its more accurate flow).
ESR scores are similar: $2.39$ for E-RAFT and $2.45$ for MultiCM.
While we focus on the inference of DNN in the proposed pipeline,
future research could look into accommodating training within the proposed joint estimation framework.

\subsection{Runtime}

Using $30$k events from the ECD dataset ($240\times180$ px) on a Mac M1 CPU (2020),
the denoising operations (scoring and ranking) take less than $0.1$~s extra computation in total, out of 2~s of the entire processing time (i.e., $5$\% computation increase with respect to the base method).
Nonetheless, accommodating learning-based approaches in our framework %
could be key to the trade-off between accuracy and speed.

\def\figWidth{0.3\linewidth}
\def\textWidth{1em}
\begin{figure}[t]
	\centering
    {\scriptsize
    \setlength{\tabcolsep}{1pt}
        \begin{tabular}{
	>{\centering\arraybackslash}m{\textWidth}
	>{\centering\arraybackslash}m{\figWidth} 
	>{\centering\arraybackslash}m{\figWidth}
	>{\centering\arraybackslash}m{\figWidth}}
		\\

            &\gframe{\includegraphics[width=\linewidth]{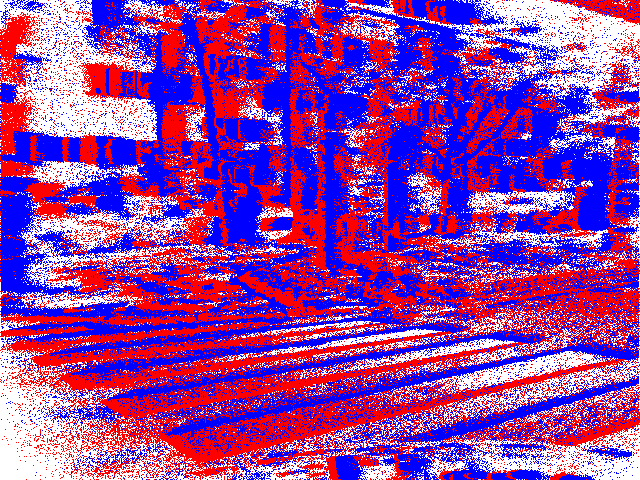}}
            &\gframe{\includegraphics[width=\linewidth]{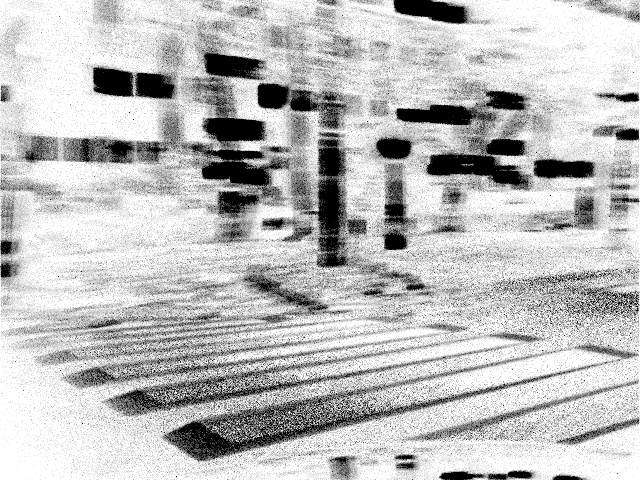}}
            \\
            & Raw Event & IWE (No Warp) & \\[0.5ex]

            \rotatebox{90}{\makecell{MultiCM \cite{Shiba22eccv}}}
            &\gframe{\includegraphics[width=\linewidth]{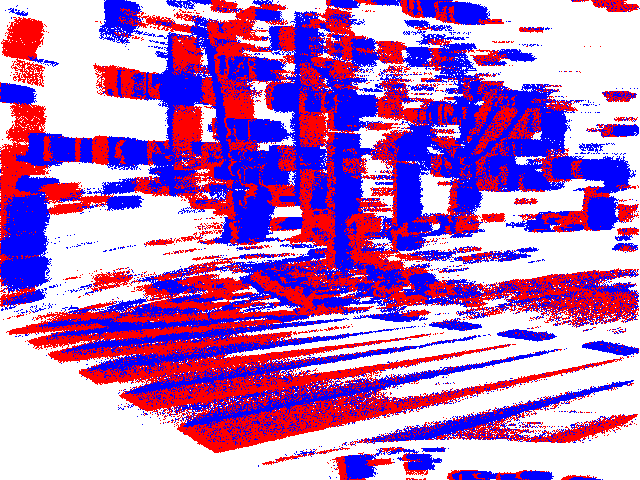}}
            &\gframe{\includegraphics[width=\linewidth]{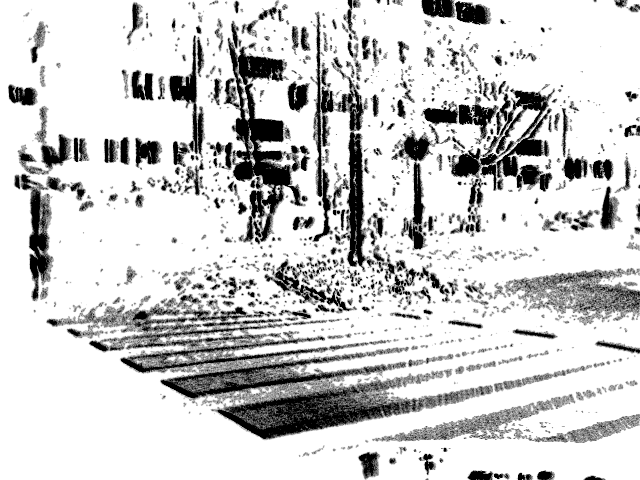}}
            &\gframe{\includegraphics[width=\linewidth]{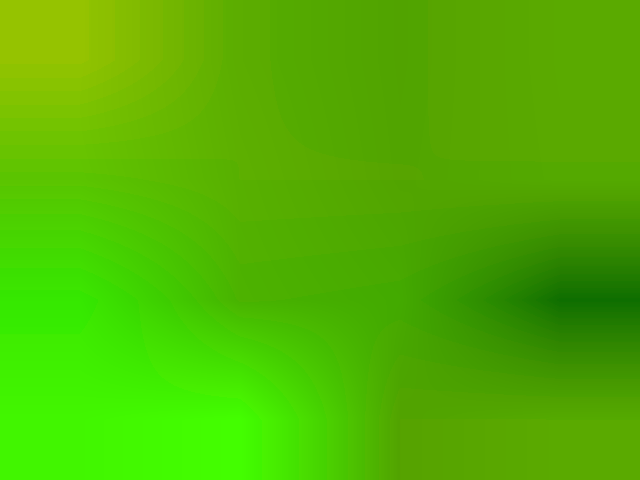}}
            \\  

            \rotatebox{90}{\makecell{E-RAFT \cite{Gehrig21threedv}}}
            &\gframe{\includegraphics[width=\linewidth]{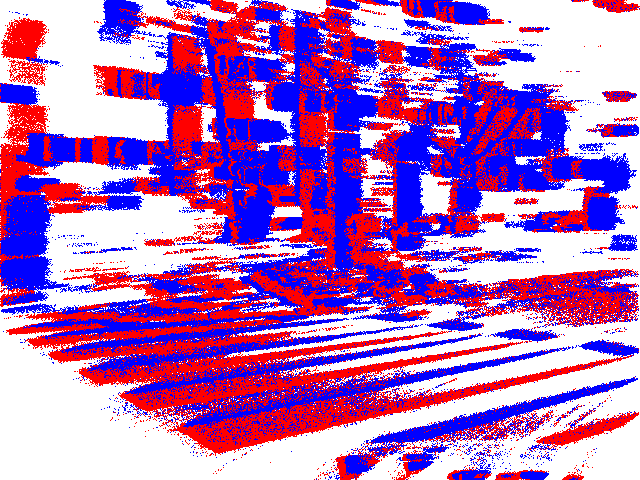}}
		      &\gframe{\includegraphics[width=\linewidth]{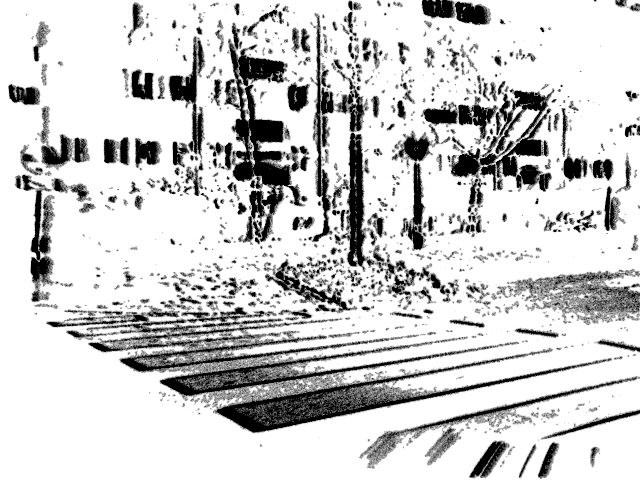}}
            & \gframe{\includegraphics[width=\linewidth]{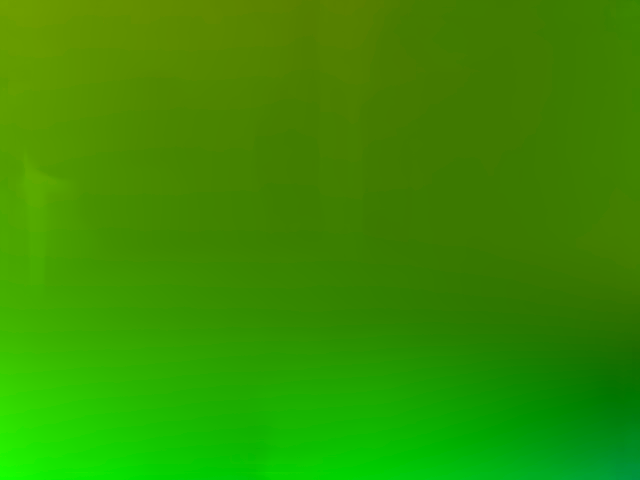}}
            \\

            & (a) Signal Events
		& (b) IWE
		& (c) Flow Estimation
		\\[-1ex]
	\end{tabular}
	}
	\caption{Effect of adopting two different motion estimators (E-RAFT or MultiCM) on the DSEC night scene (zurich\_12a) \cite{Gehrig21ral}. 
    Despite small IWE differences (e.g., in the car's hood), the proposed framework supports several motion estimators.
    }
\label{fig:result:appEraft}
\vspace{-2.3ex}
\end{figure}

\section{Limitations}
\label{sec:limitations}

Our method expects as parameter a target noise level,
but the intrinsic (``true'') noise level depends on the camera, scene, etc. and is therefore unknown. 
More research in this direction is needed for automatic parameter tuning.
Also, notice that there might be a potential limitation of the denoising metric ESR, since it depends on the motion parameter (see Suppl. Mat. of \cite{Jiang24eccv}).
Although our denoising efficacy is confirmed from multiple perspectives,
we believe that advancing denoising metrics for real-world data (where it gets interesting due to the absence of GT labels) is a central direction for future work in event-based vision.

\section{Conclusion}
\label{sec:conclusion}

In summary, this work proposes that:
($i$) Joint estimation of noise and motion is a novel problem setting compared to conventional (isolated) denoising, and is sensitive given the event data's inherent motion dependency.
($ii$) The proposed joint estimation framework can accommodate various motion estimators and has shown its efficacy in improving both denoising and motion estimation accuracies.
($iii$) It is important to evaluate denoising results on both, denoising benchmarks and downstream tasks (e.g., motion estimation, intensity reconstruction), since current denoising benchmarks have shortcomings in terms of metrics, data, and baselines.
In particular, our method achieves notable denoising results (state-of-the-art on the E-MLB benchmark and competitive on DND21) while demonstrating effectiveness across motion estimation and intensity reconstruction tasks. 
Our approach advances both event-denoising theory and practical denoising use-cases.

\section*{Acknowledgments}
We would like to thank Dr. B. Jiang for useful discussions.
Funded by
the Deutsche Forschungsgemeinschaft (DFG, German Research Foundation) under Germany’s Excellence Strategy – EXC 2002/1 ``Science of Intelligence'' – project number 390523135.

\ifarxiv

\section{Supplementary}
\label{sec:suppl}

\subsection{Video}
We encourage readers to inspect the attached video,
which summarizes the method and results.

\def\figWidth{0.31\linewidth}
\def\textWidth{1em}
\begin{figure}[h]
	\centering
    {\small
    \setlength{\tabcolsep}{1pt}

        \begin{tabular}{
	>{\centering\arraybackslash}m{\textWidth}
	>{\centering\arraybackslash}m{\figWidth} 
	>{\centering\arraybackslash}m{\figWidth}
	>{\centering\arraybackslash}m{\figWidth}}
		\\

            \rotatebox{90}{\makecell{$\text{iter} = 0$}}            
            &\gframe{\includegraphics[width=\linewidth]{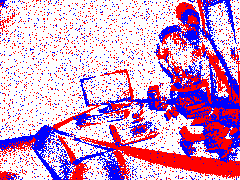}}
            &\gframe{\includegraphics[width=\linewidth]{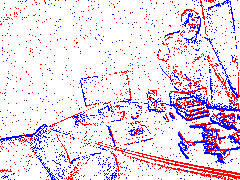}}
            &\gframe{\includegraphics[width=\linewidth]{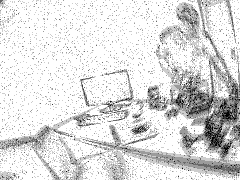}} \\

            \rotatebox{90}{\makecell{$\text{iter} = 20$}}            
            &\gframe{\includegraphics[width=\linewidth]{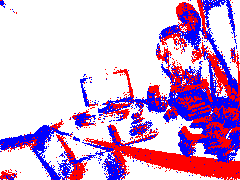}}
            &\gframe{\includegraphics[clip,trim={0.1cm 8cm 19.1cm 0cm},width=\linewidth]{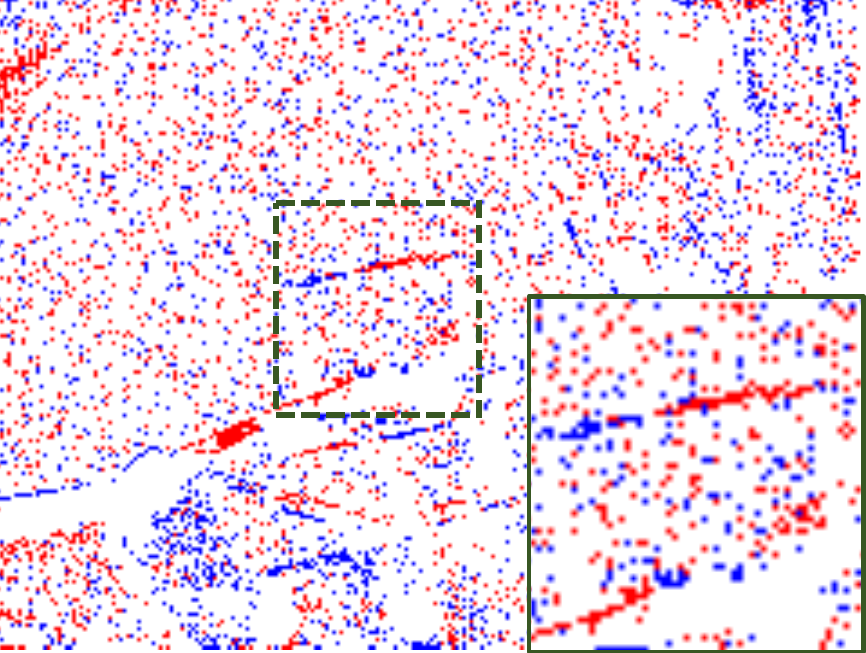}}
            &\gframe{\includegraphics[width=\linewidth]{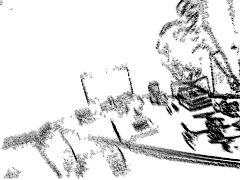}} \\

            \rotatebox{90}{\makecell{$\text{iter} = 100$}}            
            &\gframe{\includegraphics[width=\linewidth]{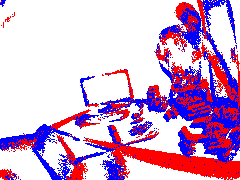}}
            &\gframe{\includegraphics[clip,trim={0.1cm 8cm 19.1cm 0cm},width=\linewidth]{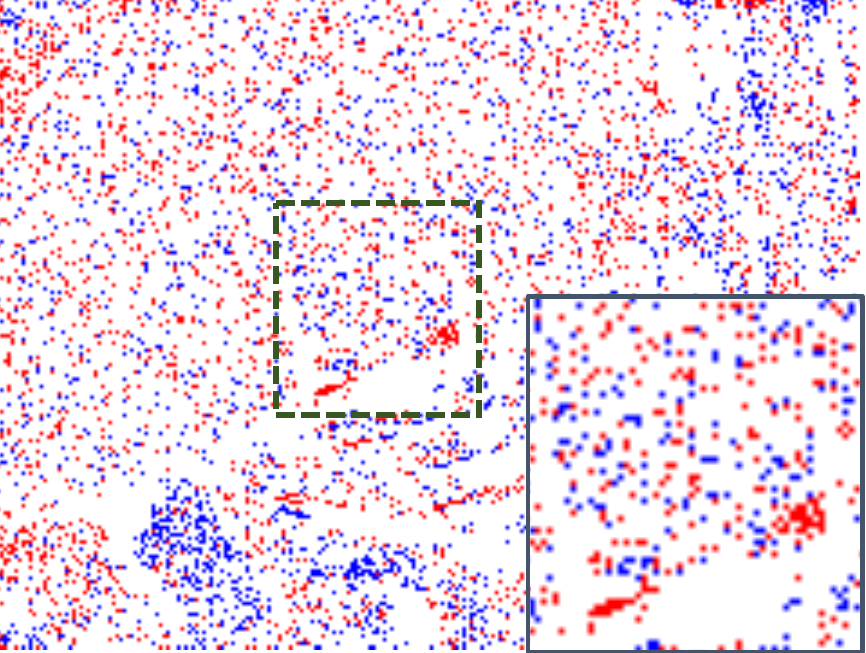}}
            &\gframe{\includegraphics[width=\linewidth]{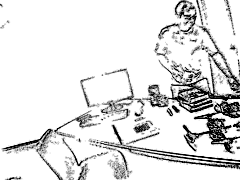}} \\
            
            & (a) Signal
            & (b) Noise
            & (c) IWE
		\\
	\end{tabular}
	}
	\caption{Evolutions of signal, noise, and motion (IWE) during optimization. The edge structure (e.g., green boxes in (b) Noise) converges to move to signal events, while CMax converges to the sharp IWE (i.e., expected motion parameters).} 
\label{fig:suppl:optim}
\end{figure}

\ifarxiv
\def\oneThirdLineWidth{0.62\linewidth}
\def\halfLineWidth{0.47\linewidth}
\def\halfLessLineWidth{0.44\linewidth}
\begin{figure}[t!]
\centering
\begin{subfigure}{\oneThirdLineWidth}
  \centering
  \gframe{\includegraphics[width=\linewidth]{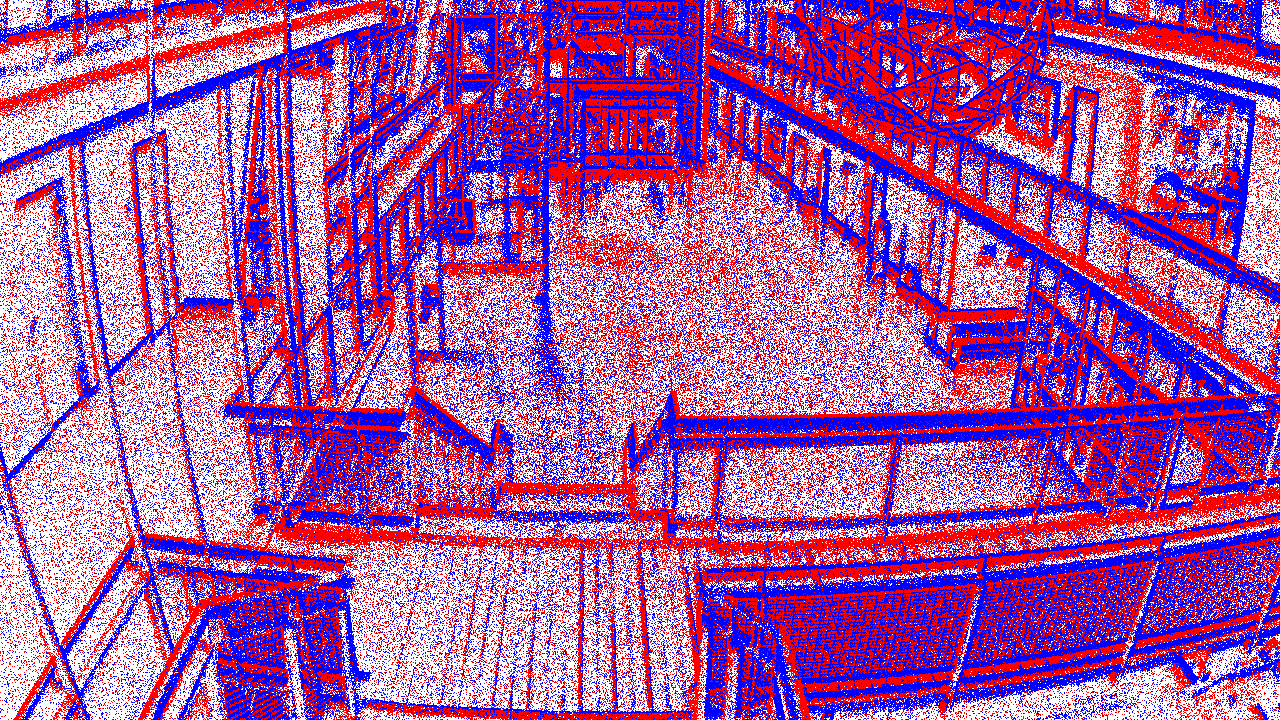}}
  \caption{Signal (first iter)}
\end{subfigure}
\begin{subfigure}{\oneThirdLineWidth}
  \centering
  \gframe{\includegraphics[width=\linewidth]{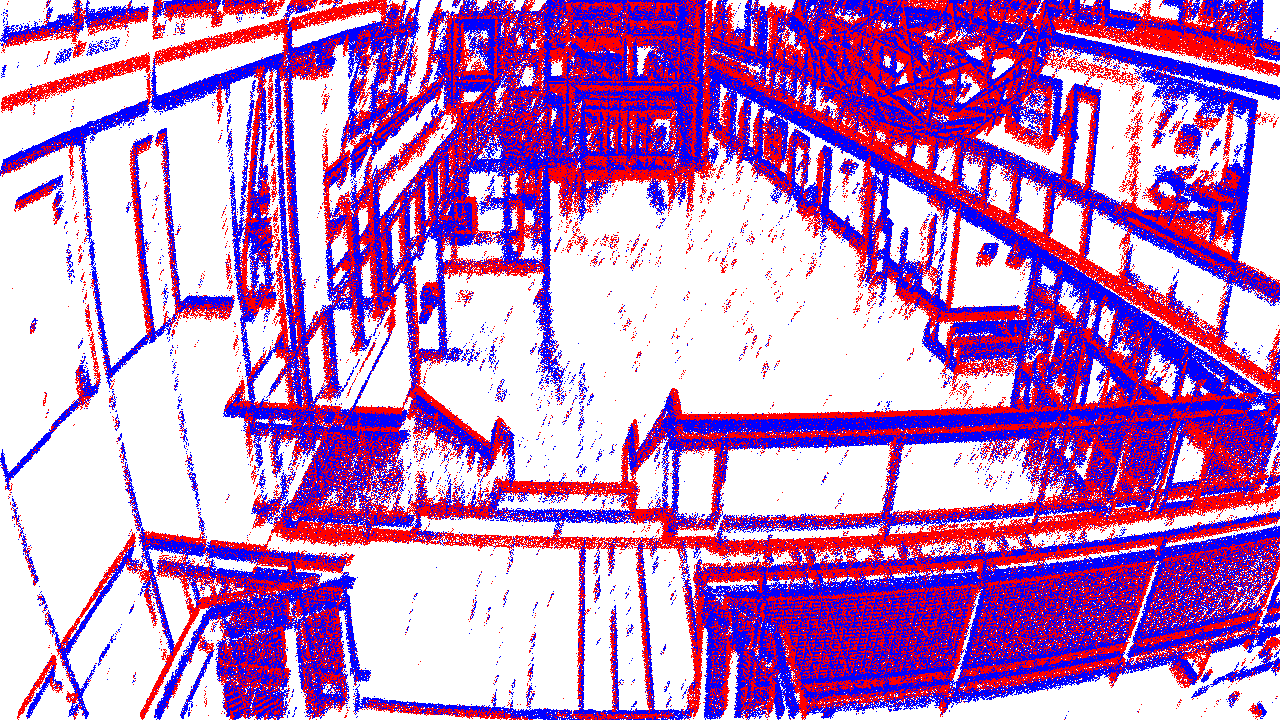}}
  \caption{Signal (last iter)}
\end{subfigure}
\begin{subfigure}{\oneThirdLineWidth}
  \centering
  \gframe{\includegraphics[width=\linewidth]{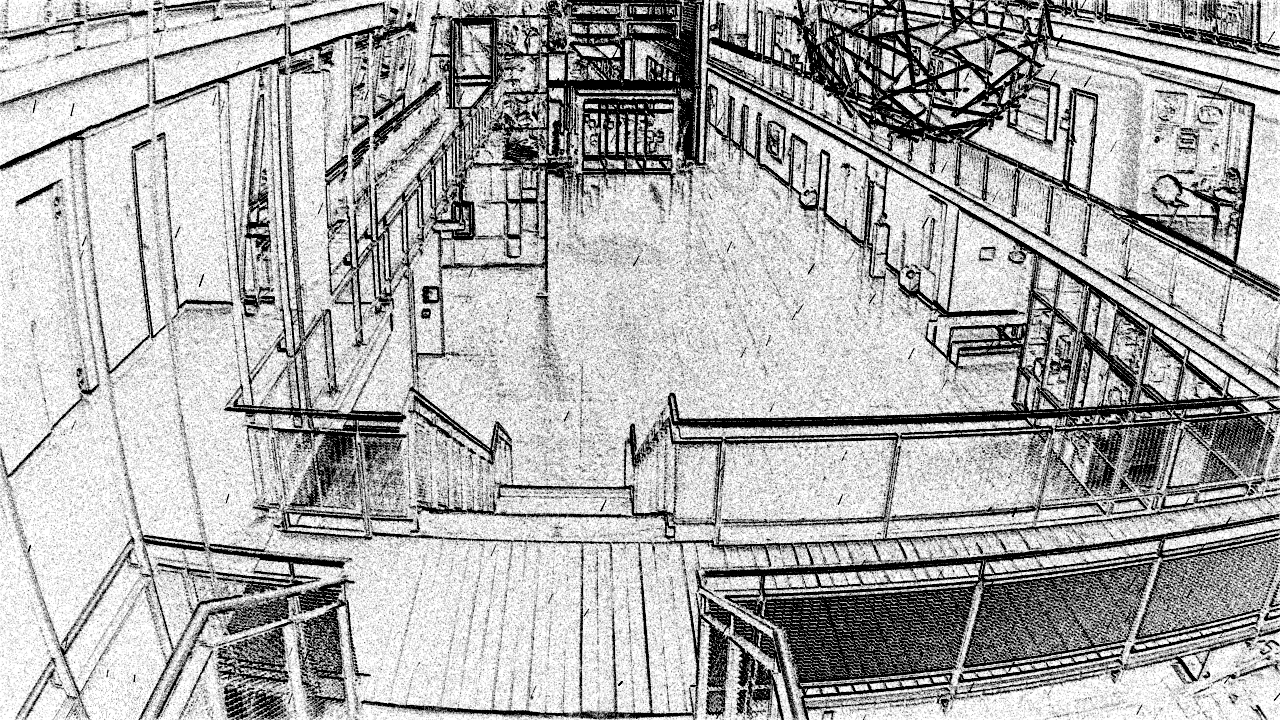}}
  \caption{IWE (last iter)}
\end{subfigure}
\\[1ex]
\begin{subfigure}{0.8\linewidth}
  \centering
  {\includegraphics[trim={0.6cm 0.8cm 0.8cm 0.6cm},clip,width=\linewidth]{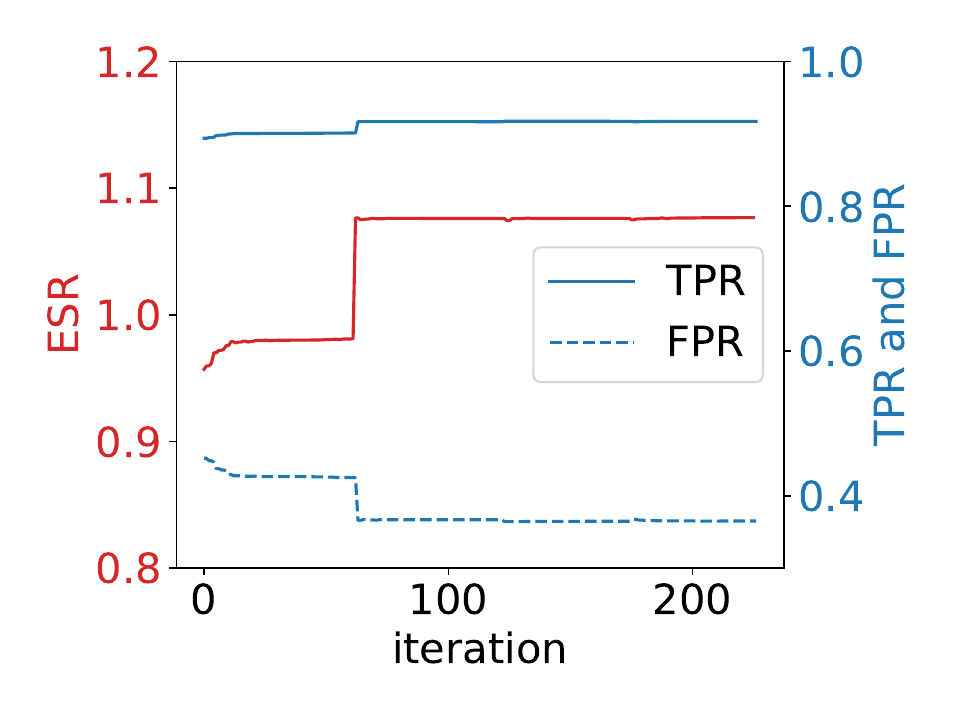}} 
    \caption{Evolution of denoising metrics}
\end{subfigure}
\\[1ex]
\begin{subfigure}{0.8\linewidth}
  \centering
  {\includegraphics[trim={0 7cm 17.1cm 0},clip,width=0.92\linewidth]{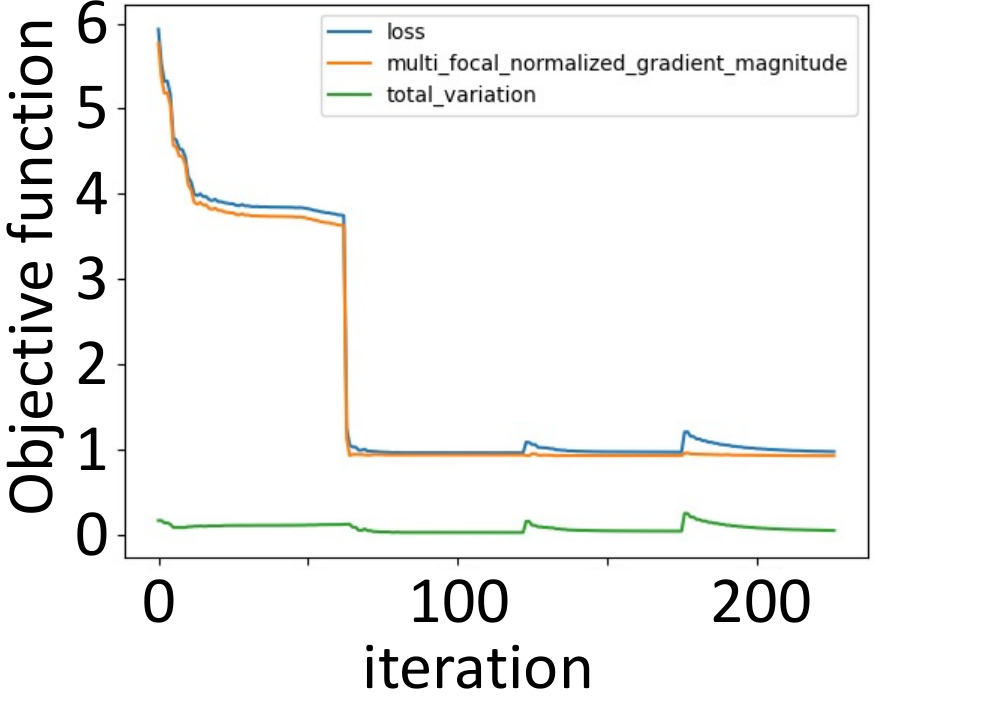}}
\caption{Evolution of loss function}
\end{subfigure}
\caption{Results on HD real-world data and intermediate denoising values (TPR, FPR and ESR metrics) during optimization.\label{fig:rebuttal:evolution}}
\end{figure}

\else 
\def\oneThirdLineWidth{0.326\linewidth}
\def\halfLineWidth{0.47\linewidth}
\def\halfLessLineWidth{0.44\linewidth}
\begin{figure}[t!]
\centering
\begin{subfigure}{\oneThirdLineWidth}
  \centering
  \gframe{\includegraphics[width=\linewidth]{images/rebuttal_evolution/bike-easy/denoise_signal_raw913.png}}
  \caption{Signal (first iter)}
\end{subfigure}
\begin{subfigure}{\oneThirdLineWidth}
  \centering
  \gframe{\includegraphics[width=\linewidth]{images/rebuttal_evolution/bike-easy/denoise_signal_raw912.png}}
  \caption{Signal (last iter)}
\end{subfigure}
\begin{subfigure}{\oneThirdLineWidth}
  \centering
  \gframe{\includegraphics[width=\linewidth]{images/rebuttal_evolution/bike-easy/warped912.png}}
  \caption{IWE (last iter)}
\end{subfigure}
\\[1ex]
\begin{subfigure}{\linewidth}
{\small
\centering
\begin{tabular}{
>{\centering\arraybackslash}m{\linewidth}}
  {\includegraphics[trim={0.6cm 0.8cm 0.8cm 0.6cm},clip,width=\linewidth]{images/rebuttal_evolution/denoising_bike-easy.pdf}} 
  \\
    (d) Evolution of denoising metrics \\[0.8ex]
  
  {\includegraphics[trim={0 7cm 17.1cm 0},clip,width=0.92\linewidth]{images/rebuttal_evolution/optimization_bike-easy.pdf}}  \\
    (e) Evolution of loss function
\end{tabular}
}
\end{subfigure}
\caption{Results on HD real-world data and intermediate denoising values (TPR, FPR and ESR metrics) during optimization.\label{fig:rebuttal:evolution}}
\end{figure}

\fi

\sisetup{round-mode=places,round-precision=3}
\begin{table*}[t!]
\centering
\adjustbox{max width=.75\linewidth}{%
\setlength{\tabcolsep}{4pt}
\begin{tabular}{ll*{10}{S[table-format=2.3]}}
\toprule
& & \multicolumn{2}{c}{1Hz}
& \multicolumn{2}{c}{3Hz}
& \multicolumn{2}{c}{5Hz}
& \multicolumn{2}{c}{7Hz}
& \multicolumn{2}{c}{10Hz}
 \\
 \cmidrule(l{1mm}r{1mm}){3-4}
 \cmidrule(l{1mm}r{1mm}){5-6}
 \cmidrule(l{1mm}r{1mm}){7-8}
 \cmidrule(l{1mm}r{1mm}){9-10}
 \cmidrule(l{1mm}r{1mm}){11-12}
& & \text{\emph{hotel}} & \text{\emph{driving}}
& \text{\emph{hotel}} & \text{\emph{driving}}
& \text{\emph{hotel}} & \text{\emph{driving}}
& \text{\emph{hotel}} & \text{\emph{driving}}
& \text{\emph{hotel}} & \text{\emph{driving}}
\\
\midrule
\multirow{6}{*}{\rotatebox{90}{\makecell{Model-based}}} 
& BAF \cite{Delbruck08issle} & 0.9535 & 0.8479 & 0.9197 & 0.8155 & 0.8916 & 0.7930 & 0.8662 & 0.7732 & 0.8366 & 0.7479 \\
& TS \cite{Lagorce17pami} & \unumr{1.3}{0.9716} & \unumr{1.3}{0.9307} & \bnum{0.9721} & \bnum{0.9260} & \unumr{1.3}{0.9606} & \bnum{0.9270} & \bnum{0.9654} & \bnum{0.9241} & \bnum{0.9620}\ & \bnum{0.9202} \\
& KNoise \cite{Khodamoradi18tetc} & 0.6773 & 0.6296 & 0.6521 & 0.6230 & 0.6703 & 0.6235 & 0.6583 & 0.6164 & 0.6413 & 0.6142 \\
& Ynoise \cite{Feng20ap} & 0.9690 & \bnum{0.9409} & 0.9517& \unumr{1.3}{0.9240} & 0.9234 & \unumr{1.3}{0.9093} & 0.9177 & \unumr{1.3}{0.8972} & 0.8987 & \unumr{1.3}{0.8800} \\
& DWF \cite{Guo22pami} & 0.9268 & 0.7409 & 0.8930 & 0.7099 & 0.8620 & 0.6901 & 0.8338 & 0.6747 & 0.7958 & 0.6563\\
& \textbf{Ours} & \bnum{1.0138021943171762} & 0.8821196931336948 & \unumr{1.3}{0.9679376335671419} & 0.8508380156237975 & \bnum{0.9625158156194444} & 0.8553984846381162 & \unumr{1.3}{0.9514067762355353} & 0.846774906545374 & \unumr{1.3}{0.9605647249996442} & 0.836273555102236 \\
\cmidrule(l{1mm}r{1mm}){1-12}
\multirow{3}{*}{\rotatebox{90}{\makecell{Learning}}} 
& EDnCNN \cite{Baldwin20cvpr} & 0.9573 & 0.8873 & 0.9371 & 0.8771 & 0.9365 & 0.8748 & 0.9254 & 0.8654 & 0.9006 & 0.874 \\
& MLPF \cite{Guo22pami} & \unumr{1.3}{0.9704} & \unumr{1.3}{0.8887} & \unumr{1.3}{0.9718} & \unumr{1.3}{0.8873} & \unumr{1.3}{0.9704} & \unumr{1.3}{0.8845} & \unumr{1.3}{0.9691} & \unumr{1.3}{0.8817} & \unumr{1.3}{0.9634} & \unumr{1.3}{0.8761} \\
& EDformer \cite{Jiang24eccv} & \bnum{0.9928} & \bnum{0.9541} & \bnum{0.9891} & \bnum{0.9472} & \bnum{0.9845} & \bnum{0.9424} & \bnum{0.9792} & \bnum{0.9343} & \bnum{0.9699} & \bnum{0.9264} \\
\bottomrule
\end{tabular}
}
\caption{\label{tab:dnd21_roc:all}The AUC$\uparrow$ of ROC on the two DND21 sequences (\emph{\emph{hotel}} and \emph{\emph{driving}}) at different noise rates.}
\end{table*}

\sisetup{round-mode=places,round-precision=3}
\begin{table}[t!]
\centering
\adjustbox{max width=.9\columnwidth}{%
\setlength{\tabcolsep}{4pt}
\begin{tabular}{ll*{4}{S[table-format=2.3]}}
\toprule
& & \multicolumn{2}{c}{\emph{dynamic\_rot}} & \multicolumn{2}{c}{\emph{boxes\_rot}}
 \\
 \cmidrule(l{1mm}r{1mm}){3-4}
 \cmidrule(l{1mm}r{1mm}){5-6}
& &\text{RMS $\downarrow$} & \text{FWL $\uparrow$} &\text{RMS $\downarrow$} & \text{FWL $\uparrow$}
\\
\midrule
\multirow{18}{*}{\begin{turn}{90}
$5$~Hz Noise
\end{turn}} 
& CMax \cite{Gallego18cvpr} & 20.00070968 & 1.258796097 & 124.641377 & 1.128529593 \\ 
& -- w/ Init. & 8.274876245 & 1.273697045 & 20.65924033 & 1.214159459 \\
& Downsampling $90$\% & 8.80770395 & 1.272907822 & 124.8686397 & 1.107758395 \\ 
& -- w/ Init. & 8.226219444 & 1.273725307 & 20.61861586 & \unumr{1.3}{1.214188455} \\
& Downsampling $80$\% & 9.964567299 & 1.270984539 & 161.8107978 & 1.060671111 \\ 
& -- w/ Init. & 8.243565368 & 1.273741842 & 20.79776956 &1.214169434\\
& Downsampling $70$\% & 14.39887891 & 1.262100344 & 184.5113834 & 1.030485545 \\ 
& -- w/ Init. & 8.231444773 & 1.273783943 & 23.67891309 & 1.21064326 \\
& Ours $90$\% & 8.521574073 & 1.273446296 & 97.58457484 & 1.14088066 \\ 
& -- w/ Init. & \unumr{1.3}{8.188607226} & \unumr{1.3}{1.273725299} & \unumr{2.3}{20.60433462} & \bnum{1.214179018} \\
& Ours $80$\% & 8.511006671 & 1.273293731 & 131.3556148 & 1.102904508 \\ 
& -- w/ Init. & \bnum{8.17003969} & \bnum{1.273790789} & 20.86156803 & 1.21409485\\
& Ours $70$\% & 9.180407823 & 1.272074788 & 163.6762405 & 1.061940985\\ 
& -- w/ Init. & 8.08626224 & 1.273578846 & 21.15111896 & 1.213969064\\
& BAF  & 19.67489305 & 1.25977249 & 125.0281451 & 1.127308902 \\ 
& -- w/ Init. & 8.253078202 & 1.27363699 & \bnum{19.5504366} & 1.21350548 \\ 
\midrule
\multirow{10}{*}{\begin{turn}{90}
$1$~Hz
\end{turn}} 
& CMax \cite{Gallego18cvpr} & 19.39519023 & 1.275866148 & 117.4397986 & 1.143852745  \\ 
& -- w/ Init. & 8.254263263 & 1.289966655 & 20.62750771 & 1.223032688 \\
& Downsampling $90$\% & 8.676346065 & 1.289459439 &  110.5683663 & 1.130495773\\ 
& -- w/ Init. & \unumr{1.3}{8.184187861} & 1.289611912 & \unumr{1.3}{20.62035821} & \unumr{1.3}{1.223050332}\\
& Ours $90$\% & 8.505807828 & \bnum{1.289787089} & 87.77454186 & 1.158678134 \\ 
& -- w/ Init. & \bnum{8.177036354} & \unumr{1.3}{1.290007185} & \bnum{20.56919161} & \bnum{1.22305286} \\
& BAF & 19.56890537 & 1.275738122 & 117.5543314 & 1.143238078 \\ 
& -- w/ Init. & 8.189429393 & 1.289532035 & 20.7134375 & 1.223006374 \\ 
\bottomrule
\end{tabular}
}
\caption{\label{tab:angvel}Angular velocity estimation on ECD dataset \cite{Mueggler17ijrr}. 
}
\vspace{-2ex}
\end{table}

\subsection{Experiment details}

\textbf{Hyper-parameters}.
For the denoising experiments,
we test various values $\snratio=\{0.9,\ldots,0.1\}$ to calculate the ROC on DND21 and $\snratio=\{0.9,\ldots,0.7\}$ for the RMS on ECD.
For E-MLB benchmarking, we fix the number of signal events to follow the prior work \cite{Ding23tom}.
To analyze the proposed pipeline in the CMax framework,
we use model-based rotational motion estimation \cite{Gallego17ral} and tile-based optical flow estimation \cite{Shiba22eccv} approaches.
We use the magnitude of the IWE gradient \cite{Gallego19cvpr} as the CMax objective function.

\subsection{Denoising Convergence During Optimization}

To further validate the proposed joint estimation approach,
we analyze the convergence during the joint estimation using the ECD dataset in \cref{fig:suppl:optim} (see also results in \cref{fig:result:angvelComp}).
In the first iteration (i.e., \emph{initialization}),
signal and noise events are randomly classified.
As optimization proceeds,
signal events evolve towards keeping edge structures in the scene,
while noise events evolve towards dropping such edge structures (see second and third rows).
Also, the IWEs converge to sharp edges with correct motion parameters.
This example confirms the efficacy of joint estimation.

The optimization process on an HD-resolution ($1280\times720$ px) real-world dataset, TUM-VIE \cite{Klenk21iros}, is shown in \cref{fig:rebuttal:evolution}.
The intermediate progress (d)--(e) demonstrates how motion estimation converges and the denoising performance improves, simultaneously.

\subsection{Full Results on Denoising DND21 data}

\gblue{\Cref{tab:dnd21_roc:all} is the full version of \cref{tab:dnd21_roc} in the main paper (including added noise at rates of 3 and 7 Hz).}

\subsection{Full Results on Angular Velocity Estimation}

While the quantitative evaluation on angular velocity estimation is summarized in \cref{fig:result:angvel_quantitative},
here, we report the detailed results with different target ratio parameters, also compared with other baselines such as BA Filter \cite{Delbruck08issle}.
The original CMax degrades due to noise, as reported in previous work (e.g., \cite{Arja23cvprw}).
The signal-to-noise (S/N) target ratio $\snratio$ affects accuracy: the closer it is to the actual value of noise injection, the better the results of the proposed method.
The amount of artificial noise injected is around $15$~\% for $5$~Hz and $3$~\% for $1$~Hz conditions.
Although the ``true'' noise level is unknown due to the original noise in the ECD sequences,
our method constantly produces better accuracy and FWL values than the baselines.
Please refer to \cref{sec:experim:angVel} for more discussions about the dependency on initialization and comparison with other baselines.
The AUCs for the conditions that we test ($\snratio=\{0.9,\ldots,0.7\}$) are $0.70$ (``Ours'') and $0.67$ (``Downsampling'').

\subsection{Quantitative Evaluation of Intensity Reconstruction}

In \cref{sec:experim:intensityRecon,sec:sensitivity} we show qualitative results of the intensity reconstruction application.
Here, we discuss possible quantitative evaluation.
The challenge of the quantitative evaluation lies in the quality of reference frames (i.e., ``GT'') in the existing dataset as shown in \cref{fig:result:denoiseQuality,fig:result:endRecon}:
the frames become underexposed or blurry due to their limited dynamic range,
when event data suffer from more BA noise (i.e., in dark scenes).

Nonetheless, we report non-reference image quality indices for different S/N ratios $\snratio$.
\Cref{fig:result:suppl:nriq} reports the scores of
Blind/Referenceless Image Spatial Quality Evaluator (BRISQUE) \cite{Mittal12tip} and
Naturalness Image Quality Evaluator (NIQE)  \cite{Mittal12spl},
using \emph{Bicycle-ND64-2} sequence (same as \cref{fig:sensitivity:recon}).
These scores indicate the perceptual quality of images, and smaller is better.
Although BRISQUE monotonically increases as the target ratio decreases (i.e., more events are removed),
NIQE scores the lowest at $\snratio=0.9$, indicating the best quality of the reconstructed image.
Although the results potentially suggest that it could estimate the ``true'' noise ratio in the data using the non-reference indices, which is useful for image reconstruction applications,
we leave further evaluation and discussion as future work.

\begin{figure}[t]
  \centering
    {{\includegraphics[clip,trim={0cm 6.5cm 10.6cm 0},width=\linewidth]{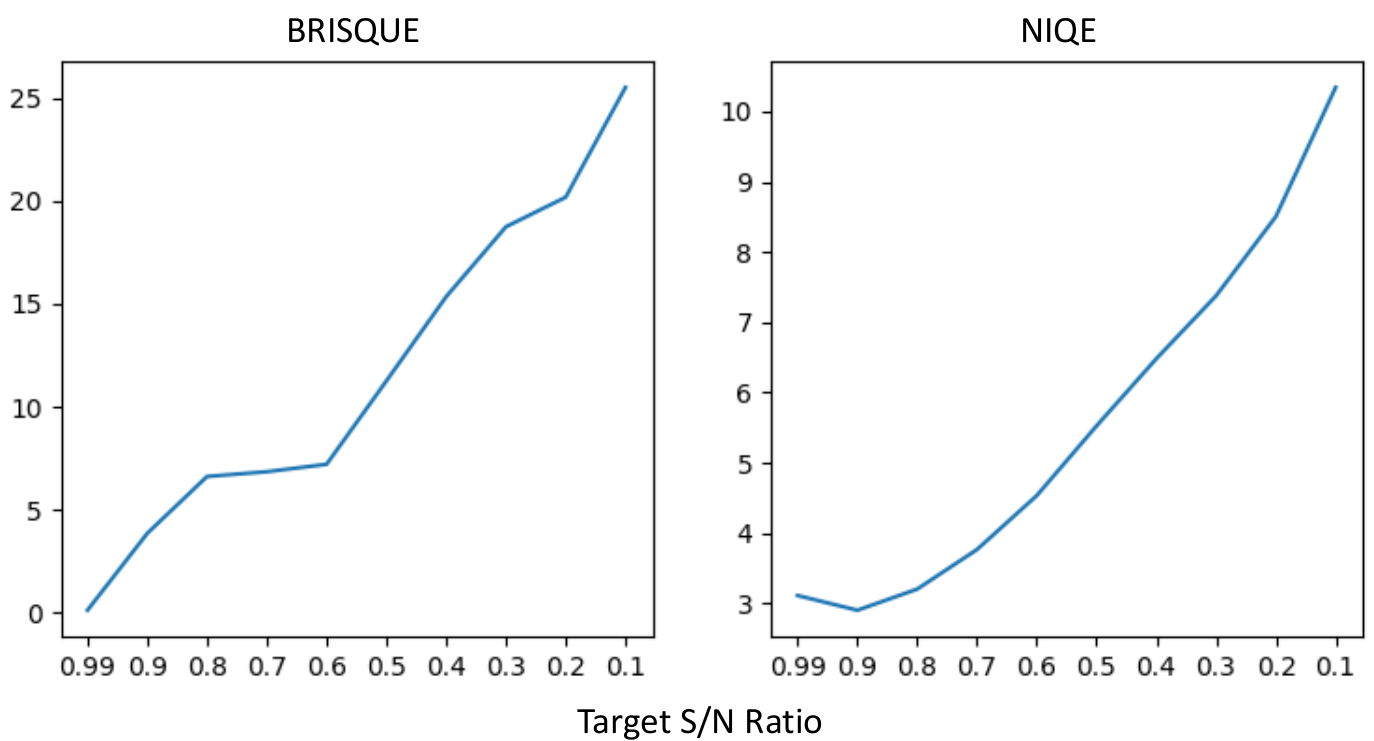}}}
    \caption{Results of the non-reference image quality indices for image reconstruction. See \cref{fig:sensitivity:recon} for details.}
    \label{fig:result:suppl:nriq}
\end{figure}

{
    \small
    \bibliographystyle{ieeenat_fullname}

}

\else 

{
    \small
    \bibliographystyle{ieeenat_fullname}
    \bibliography{all,references}
}

\cleardoublepage

\fi

\end{document}